\documentclass[10pt,preprint,3p,authoryear]{elsart_LM}
\usepackage{savesym}
\savesymbol{AND}
\usepackage{amsmath,amssymb,mathrsfs,array,stmaryrd,natbib,color,stackrel,caption,subcaption}	
\usepackage[ruled,vlined,linesnumbered]{algorithm2e}
\usepackage{empheq}

\DeclareMathAlphabet{\mathpzc}{OT1}{pzc}{m}{it}
\newtheorem{remark}{Remark}

\newcommand{\dumarg}[1]		{\widetilde{#1}}

\newcommand{\apriori}		{{\textit{a priori}}}
\newcommand{\aposteriori}	{{\textit{a posteriori}}}
\newcommand{\ie}		{\textit{i.e.}}
\newcommand{\eg}		{\textit{e.g.}}

\newcommand{\via}		{\textit{via}}
\newcommand{\etc}		{\textit{etc}}
\newcommand{\infine}		{\textit{in fine}}

\DeclareMathOperator*{\argmax}	{{\mathrm{arg\,max}}}
\DeclareMathOperator*{\argmin}	{{\mathrm{ arg\,min}}}

\DeclareMathOperator*{\dimLM}	{{\mathrm{dim}}}

\DeclareMathOperator*{\spanLM}	{{\mathrm{span}}}
\mathchardef\mhyphen="2D		

\newcommand{\ddroit}		{{\mathrm{d}}}
\newcommand{\Expesto}[1]	{\mathscr{E}_{\meas}\left[{#1}\right]}
\newcommand{\Expestoext}[1]	{\mathscr{E}_{\meas, \measnoise}\left[{#1}\right]}
\newcommand{\card}[1]		{\vert{#1}\vert}
\newcommand{\abs}[1]		{\vert{#1}\vert}

\newcommand{\ldeux}		{L^2}
\newcommand{\linf}		{L^\infty}

\newcommand{\QRequal}		{\stackrel{\rm QR}{=}}

\newcommand{\SVDequal}		{\stackrel{\rm SVD}{=}}
\newcommand{\truncSVD}		{\stackrel{\rm trunc.~SVD}{\approx}}
\newcommand{\rankapprox}[1]	{\stackrel{{\rm rank}\mhyphen{#1}}{\approx}}
\newcommand{\eigenequal}	{\stackrel{\rm eigen}{=}}
\newcommand{\IdMatrix}		{I}
\newcommand{\transpose}		{\mathsf{T}}
\newcommand{\Hermite}		{\mathsf{H}}

\newcommand{\vectorize}[1]	{\boldsymbol{#1}}
\newcommand{\matricize}[1]	{\MakeUppercase{#1}}

\newcommand{\be}	{\begin{equation}}
\newcommand{\ee}	{ \end{equation}}
\newcommand{\bi}	{\begin{itemize}}
\newcommand{\ei}	{ \end{itemize}}
\newcommand{\bea}	{\begin{eqnarray}}
\newcommand{\eea}	{ \end{eqnarray}}
\newcommand{\bc}	{\begin{center}}
\newcommand{\ec}	{ \end{center}}
\newcommand{\bealign}	{\begin{align}}
\newcommand{\eealign}	{ \end{align}}

\newcommand{\setLM}		{\mathcal{S}}
\newcommand{\R}			{\mathbb{R}}
\newcommand{\dof}		{DOFs}
\newcommand{\Gr}		{\mathrm{Gr}}
\newcommand{\approxim}[1]	{\widehat{#1}}

\newcommand{\obs}		{s}			
\newcommand{\qoi}		{y}
\newcommand{\dictionary}	{d}
\newcommand{\coef}		{x}
\newcommand{\noise}		{\eta}
\newcommand{\predictor}		{p}

\newcommand{\subobs}		{{\obs}}		
\newcommand{\subqoi}		{{\qoi}}		
\newcommand{\subsnap}		{{\mathrm{train}}}

\newcommand{\nobs}		{n_\subobs}
\newcommand{\setobs}		{\setLM_\subobs}
\newcommand{\ny}		{{n_\subqoi}}
\newcommand{\ndico}		{n_\subD}
\newcommand{\nsnap}		{n_{\subsnap}}
\newcommand{\nup}		{n_{\mathrm{up}}}

\newcommand{\qoindico}		{{{\qoi}^{\left(\ndico\right)}}}
\newcommand{\vecs}		{\vectorize{\obs}}
\newcommand{\vecsl}[1]		{\vecs_{#1}}

\newcommand{\vecy}		{\vectorize{\qoi}}
\newcommand{\vecyrestricted}[1]	{\vecy^{#1}}

\newcommand{\vecyl}[1]		{\vecy_{#1}}
\newcommand{\vecyest}		{\approxim{\vecy}}
\newcommand{\vecycentered}	{\vecy^{\circ}}		
\newcommand{\vecymean}		{\overline{\vecy}}

\newcommand{\vecycompl}		{\qoi^{\backslash \VdicoCU}}
\newcommand{\matS}		{\matricize{\obs}}
\newcommand{\matY}		{\matricize{\qoi}}
\newcommand{\matYest}		{\approxim{\matY}}
\newcommand{\matYndico}		{\matY^{\left(\ndico\right)}}
\newcommand{\vecx}		{\vectorize{\coef}}
\newcommand{\vecxest}		{\approxim{\vecx}}
\newcommand{\vecz}		{\vectorize{z}}
\newcommand{\veczsens}		{\vecz_\obs}

\newcommand{\vecxrestricted}[1]	{\vecx^{#1}}

\newcommand{\vecxlrestrdum}	{\dumarg{\vecx}_{\Ksetl}}
\newcommand{\vecw}		{\vectorize{w}}
\newcommand{\matX}		{\matricize{\coef}}
\newcommand{\vecnoise}		{\vectorize{\noise}}
\newcommand{\matB}		{{B^{\left(\dicoCU\right)}}}
\newcommand{\matBdum}		{\dumarg{B}^{\left(\dicoCU\right)}}
\newcommand{\matQ}		{Q^{\left(\matY\right)}}
\newcommand{\matR}		{R^{\left(\matY\right)}}
\newcommand{\errY}		{E^{\left(\matY,l\right)}}
\newcommand{\errR}		{E^{\left(R,l\right)}}
\newcommand{\errS}		{E^{\left(\matS,l\right)}}
\newcommand{\matRB}		{C^{\left(\dicoCU\right)}}
\newcommand{\matRBdum}		{\dumarg{C}^{\left(\dicoCU\right)}}
\newcommand{\saufl}		{{\backslash l}}

\newcommand{\vecspace}		{\mathcal{V}}
\newcommand{\Vyinf}		{\vecspace_{\qoi}}
\newcommand{\VdicoCU}		{\vecspace_{\dicoCU}}
\newcommand{\Vobs}		{\vecspace_{\vecs}}
\newcommand{\vvec}		{\vectorize{u}}

\newcommand{\dicoCU}		{\matricize{\dictionary}}
\newcommand{\dicoCUl}[1]	{\vectorize{\dictionary}_{#1}}
\newcommand{\dicoCUlinf}[1]	{\dictionary_{#1}}
\newcommand{\dicoCUrestricted}[1]	{\dicoCU_{#1}}
\newcommand{\subD}		{{\dicoCU}}

\newcommand{\dico}		{\matricize{\predictor}}
\newcommand{\dicol}[1]		{\vectorize{\predictor}_{#1}}
\newcommand{\dumdico}		{\dumarg{\dico}}
\newcommand{\obsopeinf}		{F^{\infty}}
\newcommand{\obsope}		{F}
\newcommand{\liftup}		{L}
\newcommand{\Kset}		{\mathcal{I}}
\newcommand{\Ksetl}		{\mathcal{I}^{(l)}}
\newcommand{\Sset}		{\mathcal{J}}

\newcommand{\KSVD}		{K-SVD}
\newcommand{\Ksparsity}		{K}
\newcommand{\GOBAL}		{GOBAL}

\newcommand{\QRinv}		{\textsc{QRginv}}
\newcommand{\CV}		{\mathrm{CV}}
\newcommand{\nCV}		{n_{\CV}}
\newcommand{\recerr}		{\varepsilon}

\newcommand{\spatialdomain}	{\Omega}
\newcommand{\Hilbertspace}	{\mathcal{H}}
\newcommand{\Hilbertspacefinite}{\Hilbertspace_h}
\newcommand{\meas}		{\mu}
\newcommand{\measnoise}		{\meas_{\noise}}
\newcommand{\spacephi}		{\varphi}
\newcommand{\vecspacephi}	{\vectorize{\spacephi}}
\newcommand{\eigmat}		{\varPhi}
\newcommand{\stophi}		{\psi}
\newcommand{\chronosmat}	{\varPsi}
\newcommand{\vecstophi}		{\vectorize{\stophi}}
\newcommand{\eigenvaluematrix}	{\varLambda}
\newcommand{\singvaluematrix}	{\varSigma}
\newcommand{\corrkernel}	{\kappa}
\newcommand{\empcorr}		{C}		
\newcommand{\spacevar}		{\tau}
\newcommand{\stovar}		{\theta}
\newcommand{\stovarnoise}	{\stovar_{\noise}}
\newcommand{\stodomain}		{\Theta}
\newcommand{\stomean}[1]	{\overline{#1}}
\newcommand{\stospace}		{\mathcal{W}}
\newcommand{\stospacefinite}	{\stospace_h}
\newcommand{\measspace}		{\mathcal{S}}
\newcommand{\measspacefinite}	{\measspace_h}
\newcommand{\cost}		{\mathscr{J}}

\newcommand{\ldeuxsto}		{\ldeux\left(\stodomain, \meas\right)}	
\newcommand{\ldeuxstofinite}	{\ldeux\left(\R^{\nsnap}\right)}	
\newcommand{\prodsto}[2]	{\left<{#1}, {#2}\right>_{\stospace}}	
\newcommand{\normsto}[1]	{\left\|{#1}\right\|_{\stospace}}
\newcommand{\ldeuxspacefinite}	{\ldeux\left(\R^{\ny}\right)}		
\newcommand{\prodspace}[2]	{\left<{#1}, {#2}\right>_{\Hilbertspace}}	
\newcommand{\normspace}[1]	{\left\|{#1}\right\|_{\Hilbertspace}}
\newcommand{\prodspacesto}[2]	{\left<{#1}, {#2}\right>_{\Hilbertspace \otimes \stospace}}	
\newcommand{\normspacesto}[1]	{\left\|{#1}\right\|_{\Hilbertspace \otimes \stospace}}	
\newcommand{\ldeuxmeasspace}	{\ldeux\left(\R^{\nobs}\right)}
\newcommand{\normLM}[2]		{\left\|{#1}\right\|_{#2}}
\newcommand{\normfro}[1]	{\left\|{#1}\right\|_F}

\newcommand{\probavar}		{\mathrm{p}}
\newcommand{\proba}[2]		{\probavar_{#1}\left({#2}\right)}

\newcommand{\given}		{\mid}
\newcommand{\iid}		{\textit{iid}}
\newcommand{\gauss}		{\mathcal{N}}
\newcommand{\betanoise}		{\beta}
\newcommand{\noisevariance}	{{\betanoise^{-1} \, \IdMatrix_{\nobs}}}
\newcommand{\deltanoise}	{\delta}
\newcommand{\noisevarmult}	{\deltanoise^2 \, \IdMatrix_{\nobs}}
\newcommand{\veczero}		{\boldsymbol{0}}
\newcommand{\lambdaSBL}		{\lambda}
\newcommand{\vecgamma}		{\vectorize{\gamma}}
\newcommand{\diagamma}		{\varLambda}
\newcommand{\postmeanentry}	{\mu}
\newcommand{\postmean}		{\vectorize{\postmeanentry}}
\newcommand{\postvariance}	{\varSigma}

\begin{document}

\begin{frontmatter}

\title {Observable dictionary learning for high-dimensional statistical inference}

\author[labelLIMSI]{Lionel~Mathelin\corref{cor1}}
\ead{mathelin@limsi.fr}
\author[labelSATIE,labelUJF]{K\'evin~Kasper}
\ead{kevin.kasper@univ-grenoble-alpes.fr}
\author[labelSATIE]{Hisham~Abou-Kandil}
\ead{hisham.abou-kandil@satie.ens-cachan.fr}


\address[labelLIMSI]{LIMSI-CNRS (UPR 3251), rue John von Neumann, Campus Universitaire, B\^{a}t. 508, 91405 Orsay cedex, France}
\address[labelSATIE]{Ecole Normale Sup\'erieure de Cachan, SATIE-CNRS (UMR 8029), 61 av. Pr\'esident Wilson, 94230 Cachan, France}
\address[labelUJF]{Universit\'e Grenoble Alpes, IUT1, 151 rue de la Papeterie, 38400 Saint-Martin-d'H\`eres, France}
\cortext[cor1]{Corresponding author}

\begin{abstract}

This paper introduces a method for efficiently inferring a high-dimensional distributed quantity from a few observations. The quantity of interest (QoI) is approximated in a basis (dictionary) learned from a training set. The coefficients associated with the approximation of the QoI in the basis are determined by minimizing the misfit with the observations.
To obtain a probabilistic estimate of the quantity of interest, a Bayesian approach is employed. The QoI is treated as a random field endowed with a hierarchical prior distribution so that closed-form expressions can be obtained for the posterior distribution.
The main contribution of the present work lies in the derivation of \emph{a representation basis consistent with the observation chain} used to infer the associated coefficients. The resulting dictionary is then tailored to be both observable by the sensors and accurate in approximating the posterior mean.
An algorithm for deriving such an observable dictionary is presented.
The method is illustrated with the estimation of the velocity field of an open cavity flow from a handful of wall-mounted point sensors. Comparison with standard estimation approaches relying on Principal Component Analysis and K-SVD dictionaries is provided and illustrates the superior performance of the present approach.
\end{abstract}


\end{frontmatter}

\section{Introduction}	\label{Sec_Intro}

State estimation and parameter inference are very common problems occurring in many situations. As an example, when controlling a physical system, state feedback control achieves the best results but the state vector is rarely fully mesurable and it is hence necessary to estimate it from the scarce observations (sensors) available. In another context, numerical simulations in general have now reached a degree of maturity and reliability such that the bottleneck of the accuracy of the output of the simulations typically is the quality of the input parameters. These parameters can be initial conditions, physical properties, geometry, \etc{}. Accurately estimating these input parameters is key to reliable simulations. In both the situations discussed above, the estimation of the hidden variables typically has to be done from scarce, indirect, observations of the system at hand while the information of interest is often a high-, or even infinite-, dimensional quantity. This leads to mathematically severely ill-posed problems.

A uniform grid-based description of the quantity of interest (QoI) is often employed and yields a representation under the form of a (very) high-dimensional vector, expressing a high degree of redundancy and suffering from over-parameterization. Identification of such a grid-based description through inverse techniques is a challenge.
Several approaches can be followed to address this issue. A popular one is to reduce the number of degrees of freedom to describe the quantity of interest, at the price of an approximation. If expertise knowledge is available, it can be exploited to guide the choice of the approximation space. For instance, physical quantities have finite-energy, motivating an approximation in, say, a $\ldeux$-space. Other typical examples include smoothness, known class of probability distribution, \etc{}.
This naive approach can advantageously be improved by an approach exploiting correlations of the QoI among the entries of its grid-based description. For instance, in the case the QoI is a spatial field of physical properties, as encountered, for example, in subsurface reconstruction problems, this amounts to exploiting spatial correlations and the fact that the field to be inferred typically enjoys some degree of smoothness.\footnote{Note that for physical quantities exhibiting sharp discontinuities, dedicated approaches have been proposed relying on zonation or multi-resolution, see for instance \cite{Zabaras_JCP_2016}, or multi-point statistics, \cite{Strebelle_02}.}
A representation basis is chosen and characterizing the QoI reduces to estimating the coefficients associated with the elements of the basis. This certainly is the most popular approach to inverse problems where the estimation typically reformulates as finding the small set of scalar-valued coefficients associated with the dominant modes of the approximation basis. Among many efforts from the literature, this strategy has been adopted in \cite{Buffoni2008,Yang2010} or \cite{Leroux_etal_2015}, where a Principal Component Analysis (PCA), also termed Proper Orthogonal Decomposition (POD), \cite{Holmes_etal_96,Sirovich1987}, was used as an approximation basis.
Once the approximation basis is determined, estimating the associated coefficients is done by minimizing the data misfit with the available observations. With very few observations, this data misfit minimization can itself be ill-posed and the solution method must involve some kind of regularization. A typical choice, often justified from physics-related arguments, is to look for the least energy solution, \ie{}, that with the lowest $\ldeux$-norm: the ubiquitous Tikhonov regularization. Another approach is to determine an approximation basis in which the QoI is likely to be sparse. The data misfit minimization then includes a penalization of the cardinality of the solution, in the form of a $L^0$-(quasi-)norm, and is efficiently solved using sparse coding algorithms such as Orthogonal Matching Pursuit (OMP), \cite{Mallat_Zhang_93}, CoSaMP, \cite{Needell_Tropp_08}, or Iterative Hard Thresholding (IHT), \cite{Blumensath_IHT}, to cite only a few.
For instance, this approach is followed by \cite{Jafarpour_partI_12,Jafarpour_partII_12} who have used a \KSVD{} algorithm, \cite{Aharon_KSVD,Rubinstein_etal_08}, to determine such a sparsity-promoting approximation basis in a subsurface reconstruction context.

Another strategy to address the ill-posedness of inverse problems is to explicitly introduce prior knowledge. For instance, a deterministic QoI can be treated as a random field and endowed with a prior distribution. A Bayesian approach can then be employed to ``assimilate'' the QoI from the prior to the posterior distribution using the available observations. This is a rigorous approach and results in a complete probabilistic estimation of the parameters at hand, \cite{Evans_Stark_02, Kaipio_Somersalo_05,Marzouk_JCP_09,Stuart_10}. The QoI is then described in a probabilistic sense, with closed-form expressions in some specific situations (linear forward model from the input to the observations, Gaussian distribution of the likelihood, \etc{}.) or, more generally, by sampling the input space according to the posterior measure density with a Markov-chain Monte-Carlo (MCMC) technique, \cite{Handbook_MCMC}.

Note that the two strategies can be combined and most efficient methods use both an approximation basis for the input space and a Bayesian approach. As an example, a Karhunen-Lo\`eve-based approximation of the input space, subsequently sampled with a MCMC scheme, can be used, \cite{Efendiev_etal_27,Marzouk_JCP_09}.

Besides the quality of the approximation basis or the relevance of the expert knowledge introduced as a prior in the estimation method, the performance of the recovery procedure critically depends on observations, both quantity and quality of which are key factors for an accurate inference. Among other things, including the type of acquired information, the sensors location is a critical point. In an operational situation, sensors are limited in number and are subjected to a series of constraints. For instance, in a fluid mechanics context, their are usually mounted on a solid wall, as opposed to measuring in the bulk flow. Optimal location of the sensors requires an exhaustive search and constitutes an NP-hard problem. Many approaches have been proposed to determine a reasonably good solution in a computationally acceptable way. These include Effective Independence, \cite{Kammer_91}, Optimal Driving Point, Sensor Set Expansion, Kinetic Energy Method and Variance Method, \cite{Bakir2011, Meo2005}, the maximization of the smallest singular value of the Fisher Information Matrix (FIM), \cite{Alonso2004}, compressed-sensing-inspired approaches, \cite{Bright2013,Sargsyan_etal_15,Brunton_etal_sensors}, \textsc{framesense}, \cite{Ranieri_etal_14} or the \textsc{sensorspace} method introduced by the authors of the present work, \cite{ACC_2015}.

The quality of the reconstruction from the sensor information largely depends on the \emph{observability} of the QoI from the sensors. Clearly, if the trace of the QoI on the support of the point sensors vanishes, the observations do not bear information from the QoI and the estimation is not possible. In the case of linear dynamical systems, observability of the state vector from the sensors is quantified by studying the spectrum of the observability Gramian matrix. A similar situation occurs when the QoI is estimated within an approximation basis. In this case, the quality of the estimation of the approximation coefficients largely depends on the observability of the elements of the basis from the sensors. It may happen that the sensors are well informed by some approximation modes while being blind to others. As will be seen below in our examples, this situation is quite common and strongly impacts the recovery performance.

The motivation for the present work is the following: we advocate that an efficient estimation procedure should make use of prior expertise allowing an educated choice of a suitable approximation space \emph{whose basis functions should be consistent with the way the associated coefficients will be estimated from the sensor measurements.} As will be shown below, deriving an approximation basis disregarding the way the coefficients are actually determined, as is common practice with the use of ``generic'' bases such as PCA/POD or \KSVD, may significantly affect the performance of the recovery method.

The paper is structured as follows. The general framework of inverse problems is introduced in Sec.~\ref{Sec_IP}, together with the main notations of the paper. Approximation of both an infinite- and finite-dimensional quantity is introduced and solution methods to estimate the coefficients associated with the approximation basis from observations are discussed. The need for an approximation basis aware of the sensors subsequently used to inform the coefficients is advocated and our observability-constrained basis learning approach is presented in Sec.~\ref{GOBAL_section}. Both a deterministic and a stochastic estimation methods are presented, leading to a deterministic estimate or to a full probabilistic description of the QoI when considered as a random field. The proposed basis learning method allows both cases. Computational aspects are briefly discussed and an algorithm to learn the approximation basis is presented, together with the full solution method. Our approach is illustrated with the flow field estimation from a few wall-mounted point sensors in a configuration of numerically simulated two-dimensional flow over an open cavity, Sec.~\ref{Sec_Results}. Robustness of the recovery is studied with respect to noise in the measurements. Comparisons with standard recovery techniques relying on both PCA/POD and \KSVD{} show the superior performance of our approach. The paper concludes with closing remarks in Sec.~\ref{Sec_Conclusion}.
\section{Inverse problem}	\label{Sec_IP}

\subsection{Problem statement}	\label{PB_statement}

Let us introduce the problem and associated notations. In inverse problems, one typically wants to estimate a real-valued quantity of interest $\qoi \in \Vyinf$ from a set of information $\setobs$. Few pieces of information are typically available so that the $\nobs$ real-valued elements of $\setobs$ are collected in a vector $\vecs \in \Vobs \subset \R^{\nobs}$ of \emph{observations}. The QoI can be a set of variables or a distributed quantity such as a field.
The dimension of $\Vyinf$ can be finite or infinite and we note $\ny := \dimLM\left(\Vyinf\right)$ where it is understood that $\ny$ can $\to +\infty$ in which case $\Vyinf$ is assumed to be a Hilbert space $\Hilbertspace$ of real-valued functions defined over a domain $\spatialdomain$, $\Hilbertspace \equiv \ldeux\left(\spatialdomain; \R\right)$, and endowed with an inner product $\prodspace{\cdot}{\cdot}$.

Observations are typically scarce while the representation of the field of interest involves a large number of degrees of freedom (\dof{}), so that $\nobs \ll \ny$. 

In many situations, a good approximation of $\qoi$ is sufficient for the application of interest. Further, the structure of the QoI along the coordinates $\spacevar \in \spatialdomain$, the field $\qoi\left(\spacevar\right)$ is indexed upon, can often be exploited to lower the number of \dof{} necessary for a good description. The field of interest can be approximated in a low dimensional linear subspace $\VdicoCU$ on the Grassmann manifold $\Gr\left(\ndico, \Vyinf\right)$: $\displaystyle \qoi \approx \qoindico \in \VdicoCU \subset \R^\ny$. Let $\left\{\dicoCUlinf{l}\right\}_l$ be a set of \emph{modes} defining a basis of $\VdicoCU$, the field of interest is approximated as
\be
\qoi\left(\spacevar\right) \approx \qoindico\left(\spacevar\right) = \sum_{l=1}^{\ndico} {\coef^l \, \dicoCUlinf{l}\left(\spacevar\right)}. \label{fieldapprox}
\ee

The set $\dicoCU := \left\{\dicoCUlinf{1}, \dicoCUlinf{2}, \ldots \dicoCUlinf{\ndico}\right\}$ gathers the approximation modes and is termed the \emph{dictionary}. Under this approximation, the field of interest is completely characterized in $\VdicoCU$ by the set of coefficients $\vecx \in \R^{\ndico}$.  The inverse problem then reformulates as, given a dictionary $\dicoCU$, inferring $\vecx$ from $\vecs$.

For the inference problem to admit a solution, observations $\vecs$ must be informed by the field of interest. Introducing the observation form $\obsopeinf: \qoi \in \Vyinf \mapsto \vecs \in \Vobs$, observations are related to $\qoi$ \via{} $\vecs = \obsopeinf\left(\qoi\right)$. In practice, as discussed above, the QoI is approximated in a low dimensional space $\VdicoCU$. Let $\obsope$ be a finite-dimensional continuous operator, $\obsope: \qoindico \in \VdicoCU \mapsto \vecs \in \Vobs$, the observations obey
\be
\vecs = \obsope\left(\qoindico\right) + \vecnoise, 	\label{fieldapproxlowdim}
\ee
where $\vecnoise \in \Vobs$ denotes the contribution of $\vecycompl \in \Vyinf \backslash \VdicoCU$ through $\obsopeinf$ and measurement noise, if any.

\subsection{Approximation basis}	\label{Approx_basis_POD}

To allow for a reduction of the dimensionality of the quantity to infer, an approximation basis must be chosen, see Eq.~\eqref{fieldapprox}. It should allow for an accurate approximation of $\qoi$ while being low dimensional. Prior or expertise knowledge on the QoI is typically used to guide the derivation of a suitable basis. In the context of an unknown spatially-distributed quantity over a spatial domain $\spatialdomain$, the QoI is typically considered as a statistical realization of a random field. Consider the probability space $\left(\stodomain, \sigma, \meas\right)$, where $\stodomain$ is the set of random elementary events $\stovar$, $\sigma$ the $\sigma$-algebra of the events and $\meas$ a probability measure. The space $\stospace \equiv \ldeuxsto$ of real-valued, second-order, random variables with inner product $\prodsto{\cdot}{\cdot}$ is such that
\begin{align}
\normsto{u}^2  := \prodsto{u}{u} < +\infty, \qquad
\prodsto{u}{v} := \int_{\stodomain}{u\left(\stovar\right) v\left(\stovar\right) \ddroit \meas\left(\stovar\right)}, \qquad \forall \: \left(u, v\right) \in \ldeuxsto.
\end{align}

If the QoI is a realization $\qoi \left(\spacevar; \stovar\right)$ of a second-order random field, it can be efficiently approximated in $\Hilbertspace \otimes \stospace$ using the Hilbert-Karhunen-Lo{\`e}ve decomposition, \cite{Levy_Rubinstein_99}:
\begin{align}
\qoi\left(\spacevar, \stovar\right) = \stomean{\qoi}\left(\spacevar\right) + \sum_{l = 1}^{\infty}{\lambda^{1/2}_l \spacephi_l\left(\spacevar\right) \stophi_l\left(\stovar\right)}, \qquad
\stomean{\qoi}\left(\spacevar\right) \equiv \Expesto{\qoi\left(\spacevar, \stovar\right)} := \int_{\stodomain}{\qoi\left(\spacevar, \stovar\right) \ddroit \meas\left(\stovar\right)}. \label{HKL}
\end{align}

The random variables $\left\{\stophi_l\right\}_l$ are uncorrelated in the sense of $\prodsto{\cdot}{\cdot}$, $\prodsto{\stophi_l}{\stophi_{l'}} = 0$ if $l \ne l'$, and $\left(\lambda_l, \spacephi_l\right) \in \R^+ \times \Hilbertspace$ is the ordered (decreasing $\lambda_l$) sequence of eigenvalues and eigenfunctions of the linear operator $\displaystyle \mathscr{T}_{\qoi}\left(u\right): \Hilbertspace \to \Hilbertspace := \prodspace{\corrkernel_{\qoi}}{u} = \Expesto{\left(\qoi - \stomean{\qoi} \right) \, \prodspace{\qoi - \stomean{\qoi}}{u}}$. The correlation kernel 
$\corrkernel_{\qoi} \in \Hilbertspace \otimes \Hilbertspace$ is defined as
\begin{align}
\corrkernel_{\qoi}\left(\spacevar, \spacevar'\right) := \Expesto{\left(\qoi\left(\spacevar, \stovar\right) - \stomean{\qoi}\left(\spacevar\right)\right) \, \left(\qoi\left(\spacevar', \stovar\right) - \stomean{\qoi}\left(\spacevar'\right)\right)}.
\label{corrkernel}
\end{align}

For a known kernel $\corrkernel$, a popular choice of approximation basis of a realization in $\Hilbertspace$ is the set of eigenfunctions $\left\{\spacephi_l\right\}_l$. This is justified by the classic optimality result: letting $\displaystyle \qoindico\left(\spacevar, \stovar\right) := \stomean{\qoi}\left(\spacevar\right) + \sum_{l = 1}^{\ndico}{\lambda^{1/2}_l \, \spacephi_l\left(\spacevar\right) \, \stophi_l\left(\stovar\right)}$ be the $\ndico$-term truncation of the series in Eq.~\eqref{HKL} and $\displaystyle \left\|\cdot\right\|_{\Hilbertspace \otimes \stospace}$ the norm induced by the scalar product $\displaystyle \prodspacesto{\cdot}{\cdot} := \Expesto{\prodspace{\cdot}{\cdot}}$, one infers
\begin{align}
\normspacesto{\qoi - \qoindico} = \min_{\spacephi_l \in \Hilbertspace, \, \stophi_l \in \stospace} \normspacesto{\qoi - \stomean{\qoi} - \sum_{l = 1}^{\ndico}{\lambda^{1/2}_l \, \spacephi_l \, \stophi_l}}. \label{KL_opt}
\end{align}

For a given number of terms retained, the approximation error is then of minimal norm, in the sense of Eq.~\eqref{KL_opt}. Approximating a given instance $\qoi\left(\spacevar; \stovar\right)$ of the random field in the linear $\spanLM \left\{\spacephi_l\right\}_l$ hence minimizes the \emph{average} spatial error norm $\normspace{\qoi - \qoindico}$ in the sense of $\meas$.

In practice, the correlation kernel is rarely known. However, if realizations of the random field are available, the correlation kernel can be approximated by the empirical correlation. These available realizations are usually finite-dimensional, as produced by numerical simulations or from past inference problems. A realization $\qoi\left(\spacevar;\stovar\right) \in \Vyinf$ can then be identified with $\vecy \in \R^{\ny}$ and the empirical covariance is determined from a collection of $\nsnap$ known instances of the field $\left\{\vecy_i\right\}_{i=1}^{\nsnap}$. Letting $\displaystyle \matY := \left(\vecycentered_1 \: \vecycentered_2 \ldots \vecycentered_{\nsnap} \right) \in \R^{\ny \times \nsnap}$ be the \emph{training} matrix of centered known realizations $\vecycentered_i := \vecy_i - \vecymean$, $\displaystyle \vecymean := \left(\nsnap\right)^{-1} \Sigma_{i=1}^{\nsnap}{\vecy_i}$, the empirical covariance $\empcorr_{\qoi} \in \R^{\ny \times \ny}$ is given by
\begin{align}
\empcorr_{\qoi} := \left(\nsnap - 1\right)^{-1} \: \matY \, \matY^\transpose \; \succeq \: 0 \; . \label{empcorr}
\end{align}

From the eigenvalues decomposition of the symmetric positive definite matrix $\empcorr_{\qoi} \eigenequal \eigmat \: \eigenvaluematrix \: \eigmat^\transpose$, the approximation basis can be defined by the dictionary of the eigenvectors of $\empcorr_{\qoi}$: $\dicoCU = \eigmat = \left(\vecspacephi_1 \, \vecspacephi_2 \ldots \vecspacephi_{\ny}\right)$ and the training matrix can then express as $\matY = \eigmat \, \eigenvaluematrix^{1/2} \, \chronosmat^\transpose$ with a suitable matrix $\chronosmat \in \R^{\nsnap \times \ny}$. However, one typically has a limited training set while the realizations are finely discretized so that $\nsnap < \ny$. The matrix $\empcorr_{\qoi}$ is hence not full-rank and leads to a poor approximation of the true covariance. Further, it involves the dimension $\ny$ of the discretized field, which typically is large, leading to a significant computational burden.

From now on, and without loss of generality, the data are assumed centered. With a collection of finite-dimensional (centered) training instances $\left\{\vecy_i\right\}_i$, and in the context of a second-order QoI, $\qoi \in \ldeuxspacefinite \otimes \ldeuxstofinite \equiv \Hilbertspacefinite \otimes \stospacefinite$, $\Hilbertspacefinite \subset \Vyinf$, the expression in Eq.~\eqref{KL_opt} is reformulated in
\begin{align}
\normfro{\matY - \matYndico} = \min_{\vecspacephi_l \in \Hilbertspacefinite, \, \vecstophi_l \in \stospacefinite} \normfro{\matY - \sum_{l = 1}^{\ndico}{\lambda^{1/2}_l \, \vecspacephi_l \otimes \vecstophi_l}}. \label{KL_opt_finite}
\end{align}

The Eckart-Young theorem, \cite{Eckart_Young_36}, allows to directly identify $\left\{\vecspacephi_l\right\}_l$ with the $\ndico$ dominant left singular vectors of $\matY$ and $\left\{\vecstophi_l\right\}_l$ and $\left\{\lambda_l\right\}_l$ with the corresponding right singular vectors and singular values. Letting $\singvaluematrix = \eigenvaluematrix^{1/2}$ be the diagonal $\ndico \times \ndico$ matrix of the positive square root of (positive) eigenvalues $\left\{\lambda_l\right\}_l$, the training matrix is best approximated with a rank-$\ndico$ matrix as $\matY \rankapprox{\ndico} \eigmat \, \singvaluematrix \, \chronosmat^\Hermite$. The dictionary $\dicoCU$ is then given by the dominant left singular vectors and is efficiently determined from the solution $\chronosmat$ of the following $\nsnap$-dimensional eigen problem:
\begin{align}
\left(\matY^{\transpose} \matY\right) \, \chronosmat \eigenequal \chronosmat \, \eigenvaluematrix, \qquad \dicoCU = \matY \, \chronosmat \, \eigenvaluematrix^{-1/2} = \left(\dicoCUl{1} \: \dicoCUl{2} \ldots \dicoCUl{\ndico}\right). \label{Sirovich}
\end{align}

The resulting approximation modes $\left\{\dicoCUl{l}\right\}_l$ are sometimes termed \emph{principal components} (PCA) or \emph{empirical modes} in the context of the Proper Orthogonal Decomposition (POD), \cite{Holmes_etal_96}.

\subsection{Direct approach}	\label{Direct_approach}

Once an approximation basis for the QoI is determined, the problem becomes to identify the coefficients associated with the approximation, $\displaystyle \vecy \approx \dicoCU \, \vecx$, from the only pieces of available information $\vecs$. The link between measurements $\vecs$ and the QoI $\vecy$ is similar as in Eq.~\eqref{fieldapproxlowdim} and the estimation $\vecyest$ can formulate as minimizing the $\ldeux$-norm of the data misfit:
\begin{align}
\vecy \approx \vecyest = \dicoCU \, \vecx, \qquad \mathrm{with} \qquad \vecx \in \argmin_{\dumarg{\vecx} \in \R^{\ndico}} \, \left\|\vecs - \obsope\left(\dicoCU \, \dumarg{\vecx}\right) \right\|_2^2. \label{data_misfit}
\end{align}

In the present study, $\obsope: \Hilbertspacefinite \to \ldeuxmeasspace$ is a bounded finite-dimensional linear operator and the measurements have finite energy, $\vecs \in \ldeuxmeasspace \equiv \measspacefinite \subset \Vobs$. Minimizing the data misfit then reduces to
\begin{align}
\vecx \in \argmin_{\dumarg{\vecx} \in \R^{\ndico}} \, \left\|\vecs - \obsope \, \dicoCU \, \dumarg{\vecx} \right\|_2^2. \label{data_misfit_linear}
\end{align}

Since one has limited measurements compared to the size of the dictionary, $\nobs \le \ndico$, the linear system of normal equations associated with Eq.~\eqref{data_misfit_linear} is underdetermined. Additional knowledge or constraints must hence be considered to select among the infinite number of solutions. A popular choice is to retain the minimum $\ldeux$-norm solution. This is partly justified by the heuristic that one looks for the QoI of minimal energy compatible with the observations. The inferred field is then given by
\begin{align}
\vecy \approx \vecyest = \dicoCU \, \vecx, \qquad \mathrm{with} \qquad \vecx = \left(\obsope \, \dicoCU\right)^+ \vecs, \label{pinv_sol}
\end{align}
where the superscript $^+$ denotes the Moore-Penrose pseudo-inverse. The lift-up operator $\liftup: \measspacefinite \to \Hilbertspacefinite$ from measurements to the estimated QoI is then $\liftup = \dicoCU \, \left(\obsope \, \dicoCU\right)^+$ and $\vecyest = \liftup \, \vecs$.

\subsection{Parsimonious approach}	\label{KSVD_section}

An alternative choice among the infinite number of solutions to problem \eqref{data_misfit_linear} is to select a solution involving at most $\nobs$ distinct non-zero coefficients. Upon knowledge of the corresponding set of indices $\Kset \subseteq \left\{1, 2, \ldots, \ndico\right\} \in \mathbb{N}^\ast$, $\card{\Kset} \le \nobs$, one can define the coefficient vector restricted to these non-zero coefficients: $\vecx^{\Kset} \in \R^{\card{\Kset}}$. Let $\dicoCUrestricted{\Kset} \in \R^{\ny \times \card{\Kset}}$ be the restriction of the dictionary to its columns indexed by $\Kset$, the inferred field is given by the least squares solution
\begin{align}
\vecy \approx \vecyest = \dicoCUrestricted{\Kset} \, \vecxrestricted{\Kset}, \qquad \mathrm{with} \qquad \vecxrestricted{\Kset} = \left(\left(\obsope \, \dicoCUrestricted{\Kset}\right)^\transpose \, \obsope \, \dicoCUrestricted{\Kset}\right)^{-1} \, \left(\obsope \, \dicoCUrestricted{\Kset}\right)^\transpose \, \vecs. \label{Kset_sol}
\end{align}

This approach results in a well-posed problem for $\vecx$ \emph{once} $\Kset$ \emph{is determined}.

Determining a sparse solution to a linear system of equations has attracted significant efforts in the recent past. A decisive step was made with the Compressed Sensing theory, \cite{Candes_Tao_decoding, Candes_Romberg_Tao_05, Donoho_06, Candes_Romberg_Tao_06}, which has rigorously justified the formulation of an underdetermined linear problem $A \, \vecx \approx \vectorize{b}$, $A$ being a tall $n \times p$ matrix, $n < p$, with a sparse solution as a constrained convex optimization problem (basis pursuit denoising problem, BPDN):
\begin{align}
\vecx \in \argmin_{\dumarg{\vecx} \in \R^p} \, \left\|A \, \dumarg{\vecx} - \vectorize{b}\right\|_2^2 + \rho \, \left\|\dumarg{\vecx}\right\|_1, \label{BPDN}
\end{align}
where $\rho > 0$ is a regularization parameter.

In case the cardinality of $\vecx$ is known or imposed, dedicated algorithms relying on a greedy approach, such as Orthogonal Matching Pursuit (OMP), \cite{Mallat_Zhang_93}, CoSaMP, \cite{Needell_Tropp_08} or Iterative Hard Thresholding (IHT), \cite{Blumensath_IHT}, can be employed.

\subsection{Dictionary learning}	\label{DL}

As discussed in the previous section, if the solution to a linear system is sparse, efficient solution methods exist. However, the approximation of the field of interest $\vecy$ with a sparse coefficient vector $\vecx$ associated with the dictionary $\dicoCU$ is poor, in the general case. As an example, the field of interest $\vecy$ is unlikely to admit a sparse representation in the PCA basis. The benefit of a well-posed problem for estimating the set of (sparse) coefficients is then lost by the poor resulting approximation.
This situation asks for a dictionary specifically tailored so that the associated coefficients are more likely to be sparse. That is, instead of using a generic approximation basis such as PCA, one should determine the basis\footnote{Strictly speaking, this is then not a basis since the elements $\left\{\dicoCUl{l}\right\}_l$ are not linearly independent and constitute an overcomplete set.} with the specific aim of favoring sparse solutions. Many algorithms exist to this aim and we here briefly describe the \KSVD{} method, \cite{Aharon_KSVD,Rubinstein_etal_08}, as it will be useful hereafter in the paper. For a given training set $\matY \in \R^{\ny \times \nsnap}$, the dictionary learning problem consists in finding $\dicoCU \in \R^{\ny \times \ndico}$ and $\matX \in \R^{\ndico \times \nsnap}$ such that
\begin{align}
\left\{\dicoCU, \matX\right\} \in \argmin_{\dumarg{\dicoCU}, \, \dumarg{\matX}} \, \normfro{\matY - \dumarg{\dicoCU} \, \dumarg{\matX}}^2 \qquad \mathrm{s.t.} \qquad \left\|\vecx_i\right\|_0 \le \Ksparsity, \qquad \forall \, 1 \le i \le \nsnap, \label{KSVDeq}
\end{align}
where $\Ksparsity \in \mathbb{N}^\ast$ is the maximum sparsity of the approximation for \emph{any} $\vecy_i$ from the training set.

The \KSVD{} algorithm essentially relies on an alternating minimization strategy and consists of two steps. In a Sparse Coding (SC) step, the matrix of coefficients $\matX$ is updated solving Eq.~\eqref{KSVDeq} given $\dicoCU$. While different methods can be used for this step, the original implementation of \KSVD{} relies on the OMP algorithm. By construction, each column vector of $\matX$ is, at most, $\Ksparsity$-sparse. Once the coefficients are estimated, the dictionary $\dicoCU$ is updated in a Codebook Update (CU) step, exploiting the sparse structure of $\matX$ and one iterates between these two steps until a convergence criterion is satisfied, see \cite{Aharon_KSVD} for details.

\subsection{Discussion}		\label{PODvsKSVD}

Two approximation bases have been briefly discussed above. If the correlation kernel of the QoI is known, or the empirical covariance of a representative training set can be estimated, the (Hilbert-) Karhunen-Lo\`eve decomposition can be used and leads to an approximation which is optimal in the sense of Eq.~\eqref{KL_opt_finite}. Minimization of the data misfit allows to estimate the associated vector of coefficients $\vecx$ from the measurements. This estimation method is hereafter referred to as the Principal Component Analysis-based (PCA) method.
A major limitation of this approach is that, without additional hypotheses on the solution, the estimation problem is only well-posed for a coefficient vector of size $\ndico \le \nobs$. Since measurements are typically scarce, this can severely limit the versatility of the approximation basis and the accuracy of this approach.

The additional sparsity constraint introduced in the \KSVD{} approach, and its variants, allows to use a \emph{redundant} dictionary in the sense that $\left\{\dicoCUl{l}\right\}_l$ constitutes a frame for $\Hilbertspacefinite$. This redundancy property is key as it allows to rely on dictionary with more modes than measurements while still dealing with a solvable problem for $\vecx$ thanks to the suitable sparsity constraint that one should have $\Ksparsity \le \nobs \le \ndico$. While a \KSVD{}-based dictionary is not optimal at approximating elements of the training set $\left\{\vecyl{i}\right\}_i$ for a given size $\ndico$, it however potentially allows to represent a QoI $\vecyl{i}$ with elements from a \emph{larger} dictionary, $\ndico \ge \nobs$ than with the PCA approach where, in practice, one should have $\ndico \le \nobs$. A potentially better approximation can hence be achieved.

\section{Observability-constrained dictionary learning}	\label{GOBAL_section}

\subsection{Observability issue}

While enjoying optimality in a different sense, dictionaries derived with the PCA or the \KSVD{} approaches discussed in the previous section, and more generally most dictionary learning techniques, rely on a minimization of some norm of the approximation residual of the (centered) training set, $\displaystyle \left\|\matY - \dicoCU \, \matX\right\|$. Once the dictionary is determined, the unknown QoI $\vecy$ is estimated as $\vecyest = \dicoCU \, \vecx$, with $\vecx$ inferred from the measurements $\vecs$ by minimization of the data misfit.

This is the standard approach but it nonetheless suffers an important drawback. Indeed, the dictionary is determined \apriori{} from the training \emph{fields} $\left\{\vecyl{i}\right\}_i$ solely, while the actual QoI is, \infine{}, estimated from the \emph{measurements} $\vecs$: $\vecy \approx \vecyest = \dicoCU \, \vecx\left(\vecs\right)$ where the dependance of $\vecx$ upon $\vecs$ has been explicitly written. Instead, standard dictionary learning approaches rely on a training step where a given field is implicitly approximated as $\vecy \approx \vecyest = \dicoCU \, \vecx\left(\vecy\right)$. For instance, in the CU step of \KSVD{}, modes $\left\{\dicoCUl{l}\right\}_l$ are updated given \emph{exact} coefficient vectors $\left\{\vecx_i\right\}_i$ as evaluated from the SC step. This is significantly different from the actual situation where the coefficient vector can only be evaluated from the data misfit.

From this situation, it results that the modes of the dictionary may be poorly informed by the available data in practice. 
The modes of the dictionary are determined without accounting for the observation operator $\obsope$. In particular,
the standard approach essentially consists in looking for $\dicoCU$ such that $\matY \approx \dicoCU \, \matX$ (CU-like step) with $\matX = \dicoCU^+ \matY$ (SC-like step) and it can be interpreted as looking for a $\ndico$-dimensional autoencoder $\dicoCU$ such that
\begin{align}
\matY \approx \dicoCU \, \dicoCU^+ \matY, \label{autoencoder_KSVD}
\end{align}
where the low-rank approximation is understood in the sense of \eqref{KL_opt} or \eqref{KL_opt_finite}.

Instead, a more consistent approach is to look for a $\ndico$-dimensional autoencoder $\dicoCU$ such that $\matY \approx \dicoCU \, \matX$ (CU-like step) with $\matX = \left(\obsope \, \dicoCU\right)^+ \matS$ (SC-like step), \ie{},
\begin{align}
\matY \approx \dicoCU \, \left(\obsope \, \dicoCU\right)^+ \obsope \, \matY. \label{autoencoder_GOBAL}
\end{align}

As an illustration of the potential issues of standard approaches of Sec.~\ref{Direct_approach} and \ref{KSVD_section}, consider the situation where the observation operator $\obsope$ is simply a restriction of the field of interest to a small set of points (point measurements). The vector of observations $\vecs$ is then a subset of $\vecy$ at indices $\Sset \subseteq \left\{1, 2, \ldots, \ny\right\} \in \mathbb{N}^\ast$, $\card{\Sset} = \nobs$: $\vecs = \vecyrestricted{\Sset}$. Since the dictionary is learned from $\matY$ only, it may happen that the modes vanish over the support of the sensors, so that $\obsope \, \dicoCU = \dicoCU^{\Sset} \approx \left[0\right]_{\nobs \times \ndico}$. In this situation, the coefficients $\vecx$ cannot be estimated from minimization of the data misfit in \eqref{data_misfit_linear}: the dictionary is \emph{not observable} from the sensors.

\subsection{Goal-oriented basis learning}	\label{Sec_GOBAL}

To address the limitations of standard estimation techniques discussed above, we now introduce a Goal-Oriented BAsis Learning (\GOBAL{}) solution method. In a general, possibly infinite-dimensional, setting, we look for a redundant dictionary $\eigmat\left(\spacevar\right) = \left(\spacephi_1 \, \spacephi_2 \ldots \spacephi_{\ndico}\right)$, and an estimation technique providing the function $\stophi$ from the measurements, minimizing the associated Bayes risk:
\begin{align}
\left\{\eigmat, \stophi \right\} & \in \argmin_{\dumarg{\eigmat}, \dumarg{\stophi}} \, \Expesto{\normspace{\qoi\left(\spacevar, \stovar\right) - \qoindico\left(\spacevar, \stovar; \vecs\left(\stovar\right), \dumarg{\eigmat}, \dumarg{\stophi}\right)}^2}, \nonumber \\
& = \argmin_{\dumarg{\eigmat}, \dumarg{\stophi}} \, \Expesto{\normspace{\qoi\left(\spacevar, \stovar\right) - \sum_{l=1}^{\ndico}{\dumarg{\spacephi}_l\left(\spacevar\right) \, \dumarg{\stophi}_l\left(\stovar; \vecs\left(\stovar\right)\right)}}^2}, \label{Bayes_risk_eq}
\end{align}
with $\qoi$ the centered data, $\Expesto{\qoi\left(\cdot, \stovar\right)} = 0$.

\begin{remark}
This definition of the Bayes risk implicitly assumes perfect measurements $\vecs = \obsope \, \qoi$. In practice, noise affects the measurements so that this equation does not hold anymore and Eq.~\eqref{fieldapproxlowdim} should be accounted for instead. The definition of the Bayes risk above can be readily extended to account for a random noise with a given measure. Assuming the noise is statistically independent from the realizations $\stovar$, one can introduce the probability space it belongs to, with an associated set of random elementary events $\stovarnoise$ and probability measure $\measnoise$. Accordingly extending the definition of the expectation operator, the Bayes risk is then defined as
\begin{align}
\Expestoext{\normspace{\qoi\left(\spacevar, \stovar\right) - \qoindico\left(\spacevar, \stovar; \vecs\left(\stovar, \stovarnoise\right)\right)}^2}. \label{Bayes_risk_ext}
\end{align}
\end{remark}

Restricting to the finite-dimensional framework, we let $\dicoCU \equiv \eigmat = \left(\vecspacephi_1 \, \vecspacephi_2 \ldots \vecspacephi_{\ndico}\right)$. From a collection of a finite number $\nsnap$ of independent identically distributed (\iid{}) realizations $\left\{\vecy_i\right\}_i$ and associated (noisy) measurements $\left\{\vecsl{i}\right\}_i$, problem \eqref{Bayes_risk_eq} can reformulate as looking for the dictionary that minimizes the \emph{empirical Bayes risk}, with the Bayes risk definition~\eqref{Bayes_risk_ext}:
\begin{align}
\left\{\dicoCU, \matX \right\} & \in \argmin_{\dumarg{\dicoCU}, \dumarg{\matX}} \,\normfro{\matY - \dumarg{\dicoCU} \, \dumarg{\matX}\left(\matS\right)}^2, \label{Bayes_risk_empirical}
\end{align}
where $\matY = \left(\vecy_1 \, \vecy_2 \ldots \vecy_{\nsnap}\right)$ and $\matS = \left(\vecsl{1} \, \vecsl{2} \ldots \vecsl{\nsnap}\right)$. The dictionary learning problem then admits a formulation similar to those discussed in Sec.~\ref{Sec_IP}. The coefficient vector $\vecx$ associated with the estimation $\vecyest = \dicoCU \, \vecx$ is determined by minimization of the data misfit. Since the dictionary is redundant, $\ndico \ge \nobs$, constraints have to be considered for the estimation problem to be well-posed. Similarly to the parsimonious approach of Sec.~\ref{KSVD_section}, a sparsity constraint on $\vecx$ is enforced and Eq.~\eqref{Bayes_risk_empirical} finally reformulates as:
\begin{empheq}[left=\empheqlbrace]{align}
\matX   & \in \argmin_{\dumarg{\matX} \in \R^{\ndico \times \nsnap}} \, \normfro{\matS - \obsope \, \dicoCU \, \dumarg{\matX}}^2 \qquad \mathrm{s.t.} \qquad \normLM{\vecx_i}{0} \le \nobs, \qquad \forall \, 1 \le i \le \nsnap, \label{SC_CU_naive_X} \\
\dicoCU & \in \argmin_{\dumarg{\dicoCU} \in \R^{\ny \times \ndico}} \, \normfro{\matY - \dumarg{\dicoCU} \, \matX}^2. \label{SC_CU_naive_D}
\end{empheq}
The decomposition \eqref{SC_CU_naive_D} does not impose the basis elements to be orthogonal, in contrast with other methods such as PCA. This brings more flexibility in deriving an efficient approximation basis and results in superior potential performance. The dictionary $\dicoCU$ minimizes the training field error given the coefficients $\matX$ (CU step) while the coefficients are such that they minimize the data misfit with sparsity constraints for a given dictionary (SC step).

As such, this formulation would however lead to a poor recovery procedure. Indeed, in the general case, it is unlikely that $\vecsl{i}$ admits a good $\nobs$-sparse approximation in the frame defined by $\obsope \, \dicoCU$. The naive approach of Eqs.~\eqref{SC_CU_naive_X} and \eqref{SC_CU_naive_D} would hence lead to a large residual norm in the SC step and a poor overall performance. Instead, one could explicitly introduce a sparsifying predictor operator $\dico = \left(\dicol{1} \, \dicol{2} \ldots \dicol{\ndico}\right): \R^{\ndico} \to \measspacefinite \equiv \ldeuxmeasspace$ learned from the training set of measurements such that $\vecsl{i}$ admits a good $\nobs$-sparse approximation in the linear span of $\left\{\dicol{l}\right\}_l$, $\forall \, 1 \le i \le \nsnap$. The recovery procedure then relies on a pair of dictionaries $\left\{\dico, \dicoCU\right\}$ learned from training data with

\begin{empheq}[left=\empheqlbrace]{align}
\left\{\dico, \matX\right\} & \in \argmin_{\dumdico, \dumarg{\matX}} \, \normfro{\matS - \dumdico \, \dumarg{\matX}}^2 \qquad \mathrm{s.t.} \qquad \normLM{\vecx_i}{0} \le \nobs, \qquad \forall \, 1 \le i \le \nsnap, \label{GOBAL_system_PX} \\
\dicoCU & \in \argmin_{\dumarg{\dicoCU}} \, \normfro{\matY - \dumarg{\dicoCU} \, \matX}^2. \label{GOBAL_system_D}
\end{empheq}

The general procedure discussed above is hereafter termed the \GOBAL{} method and is now discussed in more details.

\subsection{Implementation}	\label{GOBAL_implementation}

As seen above, a given $\ny$-dimensional field $\vecy_i$ is approximated in a $\ndico$-dimensional space, $\vecy_i \approx \dicoCU \, \vecx_i$, with $\ndico \le \ny$. The $\ndico$-dimensional coefficient vector $\vecx_i$ is informed by the measurements $\vecsl{i} \in \R^{\nobs}$, with $\ndico \ge \nobs$. For a well-posed problem, $\vecx_i$ is then imposed to be $\nobs$-sparse, at most. For a given predictor $\dico$, it is then the solution of a sparse coding problem
\begin{align}
\vecx_i \in \argmin_{\dumarg{\vecx} \in \R^{\ndico}} \, \normLM{\vecsl{i} - \dico \, \dumarg{\vecx}}{2}^2 \qquad \mathrm{s.t.} \qquad \normLM{\vecx_i}{0} \le \nobs, \qquad \forall \, 1 \le i \le \nsnap. \qquad {\rm \bf [Sparse~Coding]} \label{SC_step}
\end{align}

Since $\matX$ is entirely determined from $\matS$ for a given $\dico$, the solution to \eqref{GOBAL_system_D} immediately follows:
\begin{align}
\dicoCU = \matY \, \matX^+. \qquad {\rm \bf [Estimation~Codebook~Update]} \label{Est_CU_step}
\end{align}

While $\dicoCU$ as given above minimizes the Frobenius norm of the residual $\matY - \dicoCU \, \matX$ among all matrices of rank $\ndico \le \min\left(\nsnap, \ny\right)$, the Bayes risk can still be large. One should then exploit the remaining \dof{} offered by $\dico$ to improve upon this situation.

From \eqref{GOBAL_system_PX}, $\dico$ is seen to minimize $\normfro{\matS - \dico \, \dumarg{\matX}}^2$ for a given $\dumarg{\matX}$. We here adopt an approach similar to the Codebook Update step in \KSVD{}. For a given atom index $l \in \left[1, \ndico\right] \in \mathbb{N}^\ast$, let $\Ksetl$ be the set of column indices of non-vanishing coefficients, $\Ksetl = \left\{i,~1 \le i \le \nsnap;~\dumarg{\matX}^l_i \ne 0\right\}$. Let $\errY := \matY_{\Ksetl} - \dicoCU_{\saufl} \, \dumarg{\matX}_{\Ksetl}^{\saufl} \in \R^{\ny \times \card{\Ksetl}}$. The vector $\dumarg{\vecx}^l_{\Ksetl}$ can then be obtained from the rank-1 approximation of $\errY$:
\begin{align}
\errY \SVDequal \eigmat \, \singvaluematrix \, \chronosmat^\Hermite, \qquad \dumarg{\vecx}^l_{\Ksetl} \propto \chronosmat^{\Hermite, 1}. \label{Feat_CU_step1}
\end{align}

Iterating at random over the atoms $1 \le l \le \ndico$, one can estimate the coefficients $\dumarg{\matX}$ \emph{informed from the QoI recovery error} $\errY$.

As for the coefficients $\matX$ above, the predictor $\dico$, solution to \eqref{GOBAL_system_PX}, is then determined in the spirit of the \KSVD{} algorithm. For any atom index $l$, let the measurement error matrix be $\errS := \matS_{\Ksetl} - {\dico}_{\saufl} \, \matX_{\Ksetl}^{\saufl} \in \R^{\nobs \times \card{\Ksetl}}$. The predictor $\dicol{l}$ is then determined as the minimizer of $\normfro{\errS - \dicol{l} \, \dumarg{\vecx}^l_{\Ksetl}}$, with the coefficient vector $\dumarg{\vecx}^l_{\Ksetl}$ achieving a low estimation error $\normfro{\errY}$ as given by \eqref{Feat_CU_step1}. It yields
\begin{align}
\dicol{l} \propto \errS \, \left(\dumarg{\vecx}^l_{\Ksetl}\right)^\transpose, \qquad \normLM{\dicol{l}}{2} = 1. \qquad {\rm \bf [Features~Codebook~Update]} \label{Feat_CU_step2}
\end{align}

This procedure leads to a predictor $\dico$ minimizing the data misfit while accounting for associated coefficients $\matX$ achieving a low QoI estimation error.

\subsection{Computational aspects}	\label{CPU_aspects}

While the procedure described above solely relies on simple linear algebra operations, it involves potentially large scale vectors, typically the training elements $\vecy_i \in \R^\ny$. In order to improve its computational efficiency, a number of modifications to the plain procedure is now discussed.

The estimation of the dictionary $\dicoCU$ is given by the product of the training field matrix $\matY$ and the Moore-Penrose pseudo-inverse of the real-valued $\ndico \times \nsnap$ matrix of coefficients $\matX$, Eq.~\eqref{Est_CU_step}. Efficient algorithms are available to compute $\matX^+$. However, $\matX$, as given by the Sparse Coding step in \eqref{SC_step}, is sparse, \ie{}, each of its columns only has $\nobs$ non-zero entries, at most. This structure can be exploited when computing its pseudo-inverse. In this work, we rely on the \QRinv{} procedure introduced in \cite{Katsikis_etal_11} which explicitly accounts for the sparse structure of $\matX$ to speed up computations.

In the same objective of handling large matrices, it is important to notice that $\dicoCU$, as given by Eq.~\eqref{Est_CU_step}, lies in the linear span of the columns of $\matY$. Each dictionary atom $\dicoCUl{l}$ then lies in the linear span of $\left\{\vecy_i\right\}_{i=1}^{\nsnap}$.
Let consider the QR decomposition of the training matrix, $\matY \QRequal \matQ \, \matR$. Since $\dicoCUl{l} \in \spanLM \left\{\vecy_i\right\}_i$, there exists a matrix $\matB$ such that $\dicoCU = \matY \, \matB$.
The dictionary is the solution to Eq.~\eqref{GOBAL_system_D}, which reformulates as
\begin{align}
\matB \in \argmin_{\matBdum \in \R^{\nsnap \times \ndico}} \, \normfro{\matQ \, \matR - \matQ \, \matR \, \matBdum \, \matX}^2, \qquad \dicoCU = \matY \, \matB,  \label{GOBAL_system_D_QR_matB} \\
\end{align}
or, with $\matRB$ such that $\dicoCU = \matQ \, \matRB$,
\begin{align}
\matRB \in \argmin_{\matRBdum \in \R^{\nsnap \times \ndico}} \, \normfro{\matR - \matRBdum \, \matX}^2. \label{GOBAL_system_D_QR_matRB}
\end{align}

Since $\nsnap \le \ny$, the residual in Eq.~\eqref{GOBAL_system_D_QR_matRB} involves smaller matrices than in Eq.~\eqref{GOBAL_system_D} and is hence faster to evaluate. The dimension $\ny$ of $\vecy$ then entirely vanishes from the algorithm, allowing to handle QoIs of arbitrary size, provided their QR decomposition can be computed in a prior (offline) step.

\begin{remark}
The idea above is to decompose the, potentially very large, training matrix $\matY$ in a unitary matrix $\matQ$ and a smaller $\nsnap \times \nsnap$ matrix $\matR$, so that the estimation procedure involves $\matR$ only. A potentially faster alternative to the QR decomposition discussed above relies on an eigenvalue decomposition of a small $\nsnap \times \nsnap$ symmetric matrix:
\begin{align}
\matY^\transpose \matY \eigenequal V^{(\matY)} \, \left(\varSigma^{(\matY)}\right)^2 \, \left(V^{(\matY)}\right)^\transpose, \label{QR_eigen}
\end{align}
so that there exists $\matR$ such that $\matY = \matQ \, \matR$, with $\matQ$ a unitary matrix:
\begin{align}
\matQ \leftarrow \matY \, V^{(\matY)} \, \left(\varSigma^{(\matY)}\right)^{-1}, \qquad \matR \leftarrow \varSigma^{(\matY)} \, \left(V^{(\matY)}\right)^\transpose,
\end{align}
with $\varSigma^{(\matY)}$ a diagonal matrix with non-negative entries.

Disregarding the specific decomposition used, the generic terms $\matQ$ and $\matR$ will be employed in the remainder of the paper.
\end{remark}

The last implementation aspect concerns the SVD in Eq.~\eqref{Feat_CU_step1} as part of the Features Codebook Update step. The whole \GOBAL{} procedure relies on a block-coordinate descent approach, with alternate minimizations between the three steps. As such, high accuracy in not necessarily required at early steps of the iterative procedure and the SVD can safely be approximated. Further, one should notice that only the dominant (right) singular vector $\chronosmat^{\Hermite, 1}$ is required. An inexpensive approximation of the solution to Eq.~\eqref{Feat_CU_step1} is then given by a 1-iteration alternate minimization:
\begin{align}
\vecw := \errY \, \left(\vecx^l_{\Ksetl}\right)^\transpose, \qquad \dumarg{\vecx}^l_{\Ksetl} \propto \vecw^\transpose \errY. \label{SVD_fast}
\end{align}

Using the $\matY = \matQ \, \matR$ decomposition discussed above, the approximation of the solution to \eqref{Feat_CU_step1} can be derived even more efficiently:
\begin{align}
\vecw := \errR \, \left(\vecx^l_{\Ksetl}\right)^\transpose, \qquad \dumarg{\vecx}^l_{\Ksetl} \propto \vecw^\transpose \errR. \label{SVD_faster}
\end{align}

From this coefficient vector $\dumarg{\vecx}^l_{\Ksetl}$ as estimated from the residual $\errR$, the predictor $\dicol{l}$ follows from \eqref{Feat_CU_step2}.

\subsection{Sparse Bayesian Learning}	\label{Sec_SBL}

\subsubsection{A probabilistic estimate}	\label{SBL_proba_est}

The \GOBAL{} method as discussed so far allows to learn a dictionary leading to a good sparse approximation of a set of data. From given observations $\vecs$, one gets a deterministic estimation $\vecyest = \dicoCU \, \vecx\left(\vecs\right)$. Instead, it is often useful to know the uncertainty associated with an outcome, in a broad sense. It allows, for instance, to define confidence intervals for the estimated QoI and identify spatial regions where the estimation is likely to be poor, hence possibly asking for more local information, such as additional sensors. To extend the general framework of the \GOBAL{} approach introduced in the previous section, we now adopt a Bayesian viewpoint.

In the general case where the sparsity of the solution is not known beforehand, the sparse coding step given the predictor operator $\dico$ can formulate under the form of the BPDN problem, see also Eq.~\eqref{BPDN}, where the compressible coefficient vector is given by:
\begin{align}
\vecx \in \argmin_{\dumarg{\vecx} \in \R^{\ndico}} \, \left\|\vecs - \dico \, \dumarg{\vecx}\right\|_2^2 + \rho \, \left\|\dumarg{\vecx}\right\|_1, \qquad \rho > 0.\label{BPDN_SBL}
\end{align}

Observations are assumed to be affected with additive errors $\vecnoise$ independent from the prediction:
\begin{align}
\vecs\left(\stovar, \stovarnoise\right) = \dico \, \vecx\left(\stovar\right) + \vecnoise\left(\stovarnoise\right). \label{observation_model}
\end{align}

The components of the error are modeled as zero-mean \iid{} Gaussian random variables. The measurement noise is assumed to affect similarly and independently each component of $\vecs$ and the error $\vecnoise$ is modeled as a multivariate, zero-mean, Gaussian random variable with diagonal covariance matrix $\noisevariance$: $\vecnoise~\sim~\gauss\left(\veczero, \noisevariance\right)$.

The coefficient vector is modeled as a random vector endowed with a Laplace prior density:
\begin{align}
\proba{\vecx}{\vecx \given \lambdaSBL} = \frac{\lambdaSBL}{2} \, \exp\left(-\frac{\lambdaSBL}{2} \, \normLM{\vecx}{1}\right), \label{Laplace_prior}
\end{align}
with $\lambdaSBL > 0$. In a Bayesian perspective, the coefficient vector as estimated from the observations is described by its posterior density
\begin{align}
\proba{\vecx}{\vecx \given \vecs, \betanoise, \lambdaSBL} \propto \proba{\vecs}{\vecs \given \vecx, \betanoise} \, \proba{\vecx}{\vecx \given \lambdaSBL}. \label{Eq_post}
\end{align}

This Bayesian formulation provides a complete description of the coefficient vector. In particular, since $\displaystyle \argmin_{\dumarg{\vecx}} \, \left\|\vecs - \dico \, \dumarg{\vecx}\right\|_2^2 + \rho \, \left\|\dumarg{\vecx}\right\|_1 = \argmax_{\dumarg{\vecx}} \, \proba{\vecs}{\vecs \given \dumarg{\vecx}, \betanoise} \, \proba{\vecx}{\dumarg{\vecx} \given \lambdaSBL}$, with $\lambdaSBL = \rho \, \betanoise$, the solution to the BPDN problem \eqref{BPDN_SBL} is equivalently obtained as the maximum a posteriori (MAP).

\subsubsection{Hierarchical prior}

A priori values for the parameters $\betanoise$ and $\lambdaSBL$ are difficult to set and they should be learned from the data instead. We then adopt a hyperprior formulation and rely on an approach inspired from the Sparse Bayesian Learning (SBL) described in \cite{Tipping_Faul_06} and \cite{Babacan_etal_10}. In particular, while the Laplace prior \eqref{Laplace_prior} is not conjugate to the noise model \eqref{observation_model}, a suitable three-stage hierarchical prior formulation allows to derive a closed-form expression of the MAP \via{} a constructive sequential procedure.

To this end, the prior \eqref{Laplace_prior} is substituted with a product of univariate Gaussian random variables:
\begin{align}
\proba{\vecx}{\vecx \given \vecgamma} = \prod_{l=1}^{\ndico}{\gauss\left(\coef^l \given 0, \gamma_l\right)}, \label{Gaussian_prior}
\end{align}
with $\vecgamma := \left(\gamma_1 \, \gamma_2 \ldots \gamma_{\ndico}\right)$ and the variances $\gamma_l \ge 0$, $1 \le l \le \ndico$, modeled as
\begin{align}
\proba{\gamma_l}{\gamma_l \given \lambdaSBL} = \Gamma\left(\gamma_l \given 1, \lambda \slash 2\right) = \frac{\lambdaSBL}{2} \, \exp\left(- \frac{\lambdaSBL \, \gamma_l}{2}\right), \label{prior_gammaj}
\end{align}
where $\lambdaSBL \ge 0$ has a Gamma prior with parameter $\nu > 0$
\begin{align}
\proba{\lambdaSBL}{\lambdaSBL \given \nu} = \Gamma\left(\lambda \given \nu \slash 2, \nu \slash 2\right). \label{prior_lambda}
\end{align}

The hyperparameter $\lambdaSBL$ has the same role as in Eq.~\eqref{Laplace_prior} since
\begin{align}
\proba{\vecx}{\vecx \given \lambdaSBL} = \int_{\left(\R^+\right)^{\ndico}}{\proba{\vecx}{\vecx \given \vecgamma} \, \proba{\vecgamma}{\vecgamma \given \lambdaSBL} \, \ddroit \vecgamma} = \left(\lambdaSBL \slash 2\right)^{\ndico} \, \exp\left(- \lambdaSBL^{1 \slash 2} \, \sum_{l=1}^{\ndico}{\abs{\coef^l}}\right), \label{prior_3stage}
\end{align}
is of similar form as in Eq.~\eqref{Laplace_prior}. The hierarchical prior formulation then results in a Laplace prior on $\left(\vecx \given \lambdaSBL\right)$. It is important to note that $\vecgamma$ is directly responsible for the sparsity of the solution vector $\vecx$ since, for $\gamma_l \to 0$, the prior variance of $\coef^l$ tends to zero and the posterior $\proba{\coef^l}{\coef^l \given \vecs}$ will hence be zero, almost surely.

The posterior then follows a multivariate Gaussian distribution which mean and variance admit a closed-form expression, \cite{Babacan_etal_10},
\begin{align}
\vecx \sim \gauss\left(\postmean, \postvariance\right), \qquad \postmean = \betanoise \, \postvariance \, \dico^\transpose \vecs, \qquad \postvariance = \left(\betanoise \, \dico^\transpose \dico + \diagamma\right)^{-1} , \label{post_stats}
\end{align}
with $\diagamma := \text{diag}\left(1 \slash \gamma_l\right)$, $1 \le l \le \ndico$.

Hyperparameters $\left\{\betanoise, \vecgamma, \lambdaSBL, \nu\right\}$ are determined from the data $\vecs$ \via{} a type-II evidence procedure by maximizing the (log)-marginal likelihood, \ie, by integrating out $\vecx$:
\begin{align}
\log \: \proba{}{\vecs, \vecgamma, \lambdaSBL, \betanoise, \nu} = \log \int_{\R^{\ndico}}{\proba{\vecs}{\vecs \given \vecx, \betanoise} \, \proba{\vecx}{\vecx \given \vecgamma} \, \proba{\vecgamma}{\vecgamma \given \lambdaSBL} \, \proba{\lambdaSBL}{\lambdaSBL \given \nu} \, \proba{\betanoise}{\betanoise} \, \ddroit \vecx}. \label{marginal_likelihood}
\end{align}

A careful analysis of this optimization problem allows an analytical treatment and a sequential solution procedure where the sparse quantities $\postmean$ and $\postvariance$ are iteratively updated, in a memory and computationally efficient way. In particular, manipulations only act on the active set, \ie{}, predictors $\dicol{l}$ associated with a non-vanishing entry $\postmeanentry^l$ through principled sequential additions or deletions of basis element candidates.
Our implementation heavily borrows from the constructive Sparse Bayesian Learning (SBL) algorithm, \cite{Tipping_Faul_06} and \cite{Babacan_etal_10}.

Once the coefficient random vector $\vecx$ is determined, the full probability distribution $\proba{}{\vecyest \given \vecs}$ of the inferred QoI can then be estimated and leads to the probabilistic estimate
\begin{align}
\vecyest \sim \gauss\left(\dicoCU \, \postmean, \dicoCU \, \postvariance \, \dicoCU^\transpose\right). \label{post_QoI}
\end{align}

\subsection{Algorithm}	\label{Sec_algo}

The dictionary learning procedure is summarized in Algorithm \ref{Algo_GOBAL}.

\begin{algorithm}[ht!]
\DontPrintSemicolon
\caption{Sketch of the \GOBAL{} algorithm for learning observable dictionaries -- \textbf{[Offline]}}
\label{Algo_GOBAL}
  \KwIn{A training set of $\nsnap$ pairs $\left(\vecyl{i}, \vecsl{i}\right)$: $\matY$ and $\matS$.}
  \KwOut{Features dictionary $\dico$ and QoI dictionary $\dicoCU$.}
  Let $\nobs = \card{\vecsl{i}}$ for some $i$ and $\ndico$ be the size of the dictionary, chosen such that $\nobs \le \ndico \le \nsnap$. Choose $\nup \in \left[1, \ndico\right]$ and $r_\mathrm{max}$.\;
  \textbf{Initialization}: Let $\matY = \matQ \, \matR$ (QR decomposition or eigenvalue problem, see Sec.~\ref{CPU_aspects}). \;
  From the $\ndico$-truncated SVD of $\matR$, initialize $\matRB$ with the dominant left singular vectors and $\matX$ such that $\matR \truncSVD \matRB \, \matX$. Initialize $\dico = \obsope \, \matQ \, \matRB$\;
  Set $\recerr_\mathrm{best} \leftarrow \normfro{\matR - \matRB \, \matX}$ and $r \leftarrow 0$. \;
  \While{$r < r_\mathrm{max}$ and $\recerr$ does not converge}{
    $r \leftarrow r + 1$. \;
    \textbf{[Features Codebook Update]~$\rightarrow \dico$} \;
    \For{$1 \le j \le \nup$}{
    Choose $l \in \left[1, \ndico\right] \subset \mathbb{N}^\ast$ at random, \;
    $\Ksetl \leftarrow \left\{i,~1 \le i \le \nsnap;~\matX_i^l \ne 0\right\}$, \;
    Error matrices: $\errR \leftarrow \matR_{\Ksetl} - \matRB_{\saufl} \, \matX_{\Ksetl}^{\saufl}$, $\errS \leftarrow \matS_{\Ksetl} - {\dico}_{\saufl} \, \matX_{\Ksetl}^{\saufl}$, \;
    Let $\vecw = \errR \, \left(\vecx^l_{\Ksetl}\right)^\transpose$, \;	
    Fast estimation of temporary coefficients, see Eq.~\eqref{SVD_faster}: $\vecxlrestrdum \leftarrow \vecw^\transpose \errR$, \;	
    Update the features dictionary: $\dicol{l} \leftarrow \errS \, \vecxlrestrdum^\transpose$ and normalize to $\normLM{\dicol{l}}{2} = 1$. \;
    }
    \textbf{[Sparse Coding]~$\rightarrow \matX$} \;
    \For{$1 \le i \le \nsnap$~\textup{[embarrassingly parallelizable]}}{
      Use the SBL algorithm to evaluate $\vecx_i$ from $\vecsl{i}$ given $\dico$, see Sec.~\ref{Sec_SBL}. \; Alternatively, use OMP with sparsity constraint $\Ksparsity = \nobs$. \; \label{algo_SC}
      }
    Evaluate $\recerr = \normfro{\matR - \matRB \, \matX}$. \;
    \If{$\recerr < \recerr_\mathrm{best}$}{$\recerr_\mathrm{best} \leftarrow \recerr$, $\dico_\mathrm{best} \leftarrow \dico$, $\matRB_\mathrm{best} \leftarrow \matRB$.}
    \textbf{[Estimation Codebook Update]~$\rightarrow \matRB$} \;
    Update the dictionary: $\matB \leftarrow \matX^+$~~[possibly with \QRinv{}] and $\matRB \leftarrow \matR \, \matB$.\;
    }
  Learned dictionaries are finally $\dico \leftarrow \dico_\mathrm{best}$, $\matRB \leftarrow \matRB_\mathrm{best}$. \;
  If desired, assemble the physical QoI dictionary: $\dicoCU \leftarrow \matQ \, \matRB$.
\end{algorithm}

In the online stage, \ie{}, once \textit{in situ} with only some measurements $\vecs$ are available to estimate the QoI $\vecy$, dictionaries $\dico$ and $\dicoCU$ learned in the offline (training) stage are used. The Sparse Bayesian Learning (SBL) procedure described in Sec.~\ref{Sec_SBL} can be employed to estimate the posterior mean $\postmean$ and associated covariance $\postvariance$ of the coefficients $\vecx$ associated with the atoms of the dictionary $\dicoCU$. The estimate $\vecyest\left(\vecs; \dico, \dicoCU\right)$ is finally given as in Eq.~\eqref{post_QoI} in the form of a random field.

\begin{remark}
If one is not interested in the statistics associated with the estimation, the (deterministic) coefficients $\vecx\left(\vecs; \dico\right)$ can be more directly estimated as the solution to the Sparse Coding problem, Eq.~\eqref{SC_step}, typically obtained with an Orthogonal Matching Pursuit (OMP) algorithm. The then deterministic estimate is hence $\vecyest\left(\vecs; \dico, \dicoCU\right) = \dicoCU \, \vecx$.
\end{remark}

\subsection{Numerical complexity} \label{Sec_complexity}

The complexity of the \GOBAL{} method, in its deterministic version, is now discussed.

We assume that the action of a $m \times n$ matrix on a vector is of complexity $2 \, m \, n$ and denote the number of iterations of the algorithm by $r$. We let $\nup \leftarrow \ndico$.

First, the training set matrix $\matY$ is decomposed into $\matQ$ and $\matR$ \via{} the eigenvalue decomposition of $\matY^\transpose \, \matY$, see Eq.~\eqref{QR_eigen}. The cost associated with forming the symmetric matrix is $\ny \, \nsnap^2$, while the eigenvalue decomposition takes $\mathcal{O}\left(\ndico \, \nsnap^2\right)$ operations. The cost of the initial decomposition of the training set is then
\begin{align}
\cost_\mathrm{decomp} = \ny \, \nsnap^2 + \mathcal{O}\left(\ndico \, \nsnap^2\right).
\end{align}

In the Estimation Codebook Update step, the numerical complexity associated with computing the pseudo-inverse of the full row-rank $\ndico \times \nsnap$-matrix $\matX$ is:
\begin{align}
\cost_\mathrm{ECU} = \frac{3}{2} \, \ndico^2 \, \nsnap + \frac{1}{3} \, \ndico^3,
\end{align}
at leading order, \cite{Moller1999}. It is cubic in the small number of atoms and only linear in the number of training elements, hence allowing a fast evaluation. Note that this is the complexity of the plain vanilla pseudo-inverse and that the used \QRinv{} algorithm has a lower complexity and enjoys superior computational performance.

Relying on a Cholesky-based variant of the OMP algorithm for the Sparse Coding step, \cite{Cotter_etal_99,Blumensath_Davies_08}, the number of operations is found to be, \cite{Rubinstein_etal_08},
\begin{align}
\cost_\mathrm{SC} = 2 \, \nobs \, \ndico \, \nsnap + 2 \, \nobs^2 \, \nsnap + 2 \, \nobs \left(\ndico + \nsnap\right) + \nobs^3.
\end{align}

Matrix operations involved in the Features Codebook Update step are of similar nature to those of the approximate K-SVD discussed in \cite{Rubinstein_etal_08}, so that one can borrow from their analysis. The number of operations per signal is here:
\begin{align}
\cost_\mathrm{FCU} = 4 \, \nobs \, \nsnap^2 + 6 \, \nsnap \, \nobs \, \ndico + 2 \, \nsnap \, \nobs^2 + 2 \, \ndico^2 \, \left(\nobs + \nsnap\right).
\end{align}

The framework of this study is limited observations compared to the size of training set. Further, the size of the dictionary is a small multiple $m \in \mathbb{N}^\ast$ of the number of sensors. One then has $\nobs \le \ndico \le \nsnap \le \ny$ and the total cost $\cost$ approximates as
\begin{align}
\cost & = \cost_\mathrm{decomp} + r \, \left(\cost_\mathrm{ECU} + \nsnap \, \cost_\mathrm{SC} + \cost_\mathrm{FCU}\right), \nonumber \\
& \approx \nsnap^2 \, \left(\ny + 2 \, r \, \nobs^2 \, \left(m + 1\right)\right) + \mathcal{O}\left(\nobs \, \nsnap^2\right).
\end{align}

Initialization of the training set aside (line 2 of the algorithm), the complexity of our core dictionary learning procedure is then seen to mainly scale with the square of the number of sensors and the square of the size of the training set.
The bottleneck of the numerical complexity is the Sparse Coding step, which scales with $\nsnap^2$. Note however that this step is embarrassingly parallelizable among the $\nsnap$ different elements of the training set, so that the walltime cost of this step, carried-out over $\nsnap$ cores, only scales with $\nsnap$.

For a very large training set, alternatives to the present approach can be derived, relying on an online, or mini-batch, treatment of the training set, \eg, inspired from the work of \cite{Mairal_etal_10}.

\section{Numerical experiments}	\label{Sec_Results}

\newcommand{\ratiolargeuronecol}	{0.6}
\newcommand{\ratiolargeurtwocol}	{0.5}
\newcommand{\ratiolargeurthreecol}	{0.33}
\newcommand{\ratiolargeurfourcol}	{0.245}
\newcommand{\recerror}			{\varepsilon}

\subsection{Configuration}	\label{Sec_cav_config}

The methodology introduced above is now demonstrated on a numerical simulation of the two-dimensional incompressible flow over an open cavity using our in-house code \textsc{olorin}. The cavity aspect ratio is 2 (deep cavity). The numerical grid comprises $296 \times 128$ nodes on a structured, non-uniform, staggered mesh. The flow is from left to right on the plots and a laminar boundary layer flow profile is prescribed at the entrance of the numerical domain. The flow Reynolds number $Re$, based on the cavity length and the uniform flow rate velocity, is $Re = 7500$. The Navier-Stokes equations are solved using a prediction-projection method and a finite-volume second-order centered scheme in space. Viscous terms are evaluated implicitly while convective fluxes are estimated with a linear Adams-Bashford scheme. Details about the numerical settings may be found in \cite{Rizi_etal_14,Podvin_etal_06}.

In the aim of being as realistic as possible, the retained sensors mimic actual devices routinely used in the flow control community. In particular, the sensors should be physically mounted on a solid wall and hence cannot be located in the bulk flow. They are here modeled as noisy point devices measuring the local wall shear stress, \ie{}:
\begin{align}
\obs^j\left(\veczsens^{(j)}\right) \propto \left. \frac{\partial \left\|\vvec\left(\vecz\right)\right\|_2}{\partial n}\right|_{\veczsens^{(j)}}, \qquad 1 \le j \le \nobs,
\end{align}
with $\veczsens^{(j)}$ the location of the $j$-th sensor $\vvec\left(\vecz\right)$ the fluid velocity vector and $n$ the distance to the wall along the local normal direction.

Sensor placement is of critical importance for the efficiency of any recovery procedure and is a well known problem. Many methods have been introduced to address this issue, as briefly discussed in Sec.~\ref{Sec_Intro}. Here, no sensor placement method is employed. In an attempt to convincingly demonstrate the performance of our \GOBAL{} method against alternatives approaches such as PCA and \KSVD{}, the sensors all these methods rely on are selected such that \emph{the recovery performance of \KSVD{} is maximized}. Sets of $\nobs = 10$ sensors locations $\left\{\veczsens^{(j)}\right\}_{j=1}^{\nobs}$ are sampled uniformly at random from the geometry of the cavity wall. For each realization of the sensor set, the recovery performance of the \KSVD-based approach is evaluated. The sensor location finally retained is the one leading to the best recovery performance. The performance of the \KSVD-based approach is thus significantly favored by this choice.

The objective of the estimation procedure is to recover the field of velocity magnitude. The training set comprises $\nsnap = 1000$ instances, $\left\{\vecsl{i}, \vecyl{i}\right\}_{i=1}^{\nsnap}$, and the recovery performance is quantified in terms of relative error norm:
\begin{align}
\recerr := \left.\normfro{\matY^{\CV} - \matYest^{\CV}\left(\matS^{\CV}\right)} \: \normfro{\matY^{\CV}}^{-1}\right.,
\end{align}
with $\left\{\vecsl{i}^{\CV}, \vecyl{i}^{\CV}\right\}_{i=1}^{\nCV}$ a set of $\nCV = 246$ instances not contained in the training set.

Unless explicitly specified, the \GOBAL{}-based estimate is the posterior mean.

\subsection{Dictionaries}	\label{plot_dicos}

In addition to our method \GOBAL{}, the performance of a recovery using both a PCA- and a \KSVD{}-based dictionary is studied. 
The PCA dictionary is given by the set of $\ndico$ dominant eigenvectors of the empirical correlation matrix $\matY^\transpose \matY$, see Eq.~\eqref{Sirovich}, while the \KSVD{} dictionary is determined as the solution of Eq.~\eqref{KSVDeq}, with $\Ksparsity = \nobs$ since this is the maximum number of atoms which can be informed by the available measurements. With either of these dictionaries, determined solely from $\matY$, and not from $\left\{\matS, \matY\right\}$, the estimation is given by $\vecyest_{i} = \dicoCU \, \vecxest_{i}$, $1 \le i \le \nCV$, with $\vecxest_{i}$ minimizing the data misfit in the sense of Eq.~\eqref{data_misfit_linear} (PCA approach) or \via{} a Sparse Coding step, Eq.~\eqref{SC_CU_naive_X} (\KSVD{} approach).

Since \KSVD{} and \GOBAL{} are iterative procedures, the PCA-based dictionary is used as initial condition in both these methods. Note that the mean field $\vecymean$ has been subtracted from the training set so that dictionaries are learned from centered data.

\begin{figure}[t!]
    \centering
    \begin{subfigure}[t]{\the\numexpr\ratiolargeurthreecol \textwidth}
        \centering
        \includegraphics[width=\textwidth,angle=0]{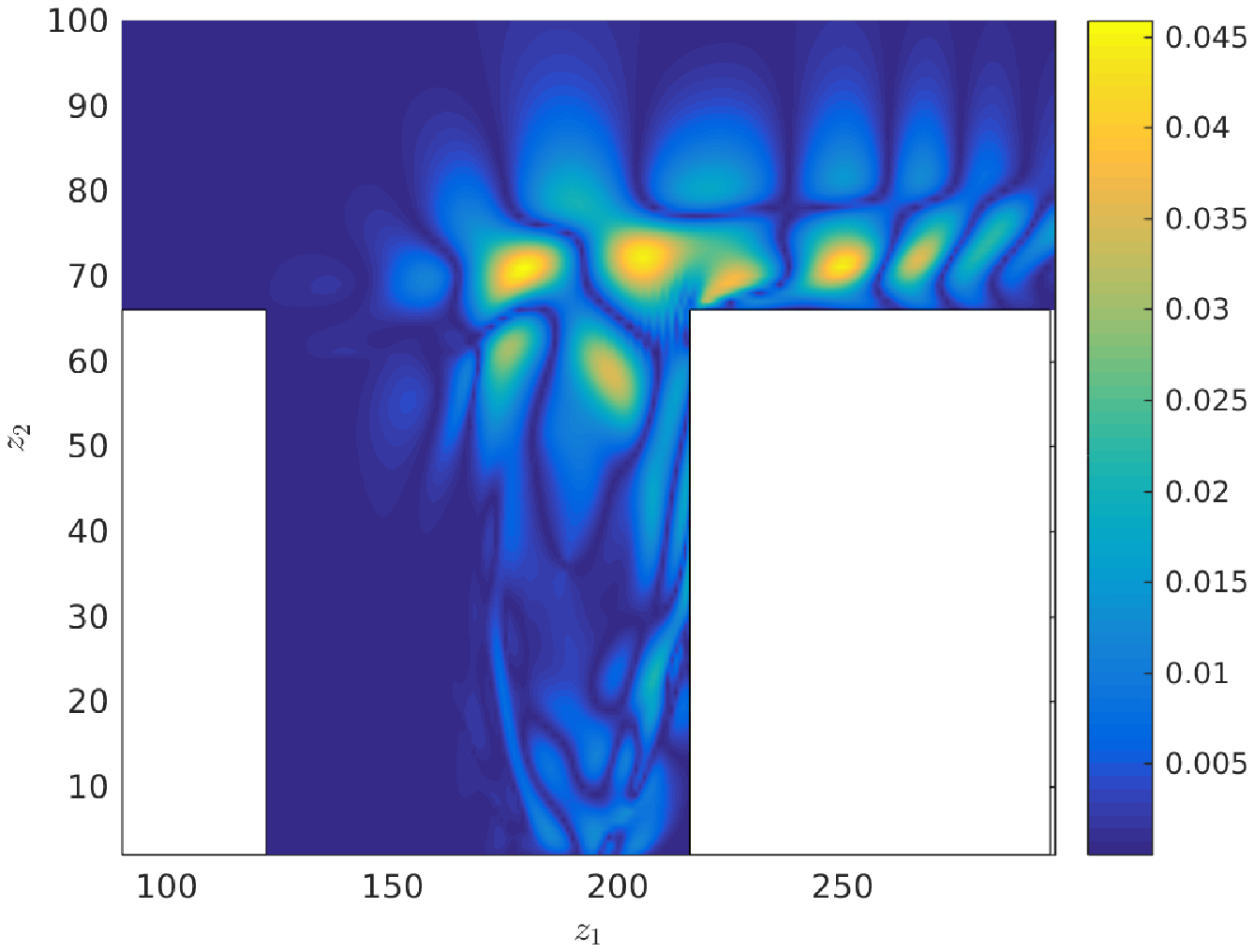}
        \caption{PCA dictionary -- Mode 1.}
    \end{subfigure}%
    \begin{subfigure}[t]{\the\numexpr\ratiolargeurthreecol \textwidth}
        \centering
        \includegraphics[width=\textwidth,angle=0]{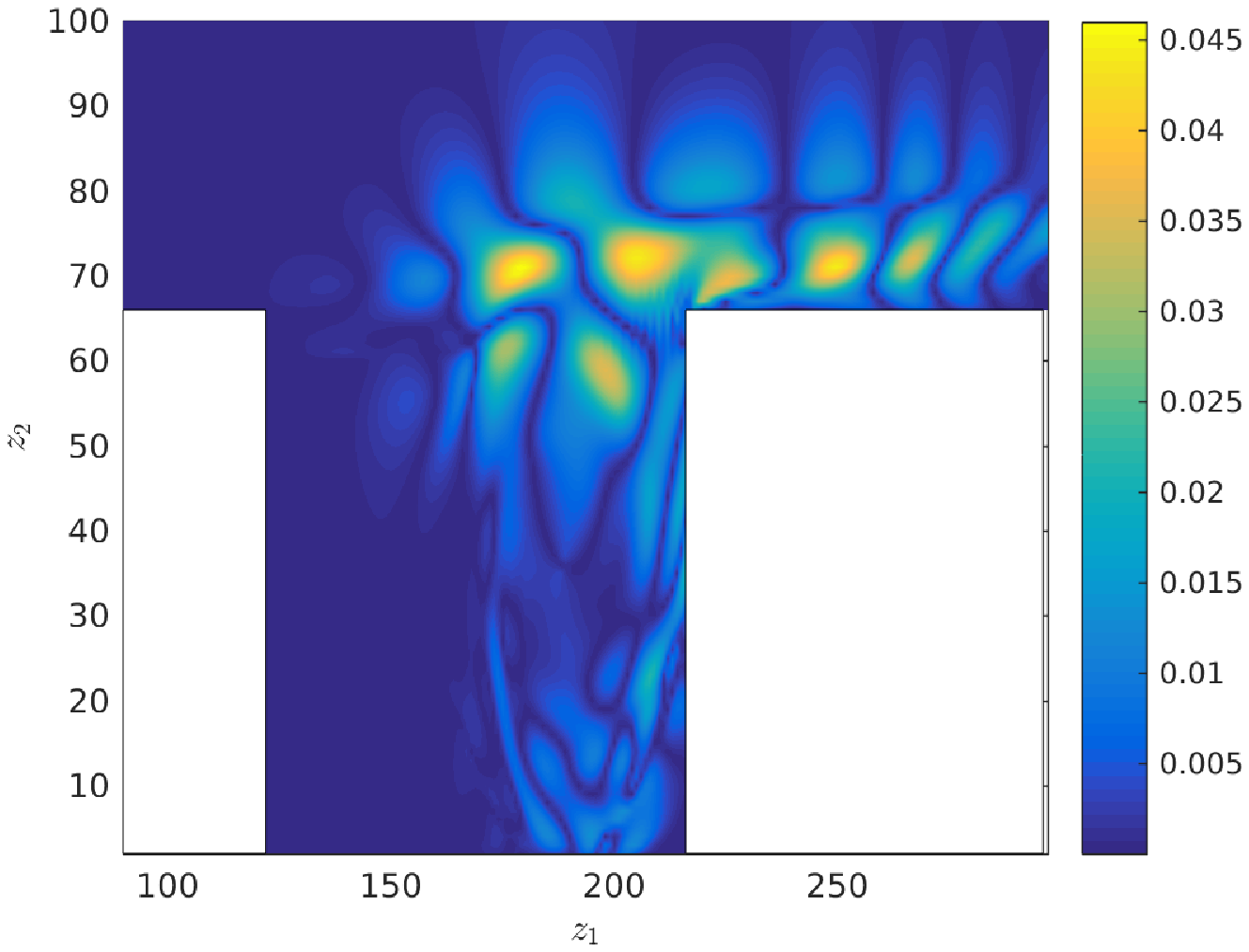}
        \caption{\KSVD{} dictionary -- Mode 1.}
    \end{subfigure}
    \begin{subfigure}[t]{\the\numexpr\ratiolargeurthreecol \textwidth}
        \centering
        \includegraphics[width=\textwidth,angle=0]{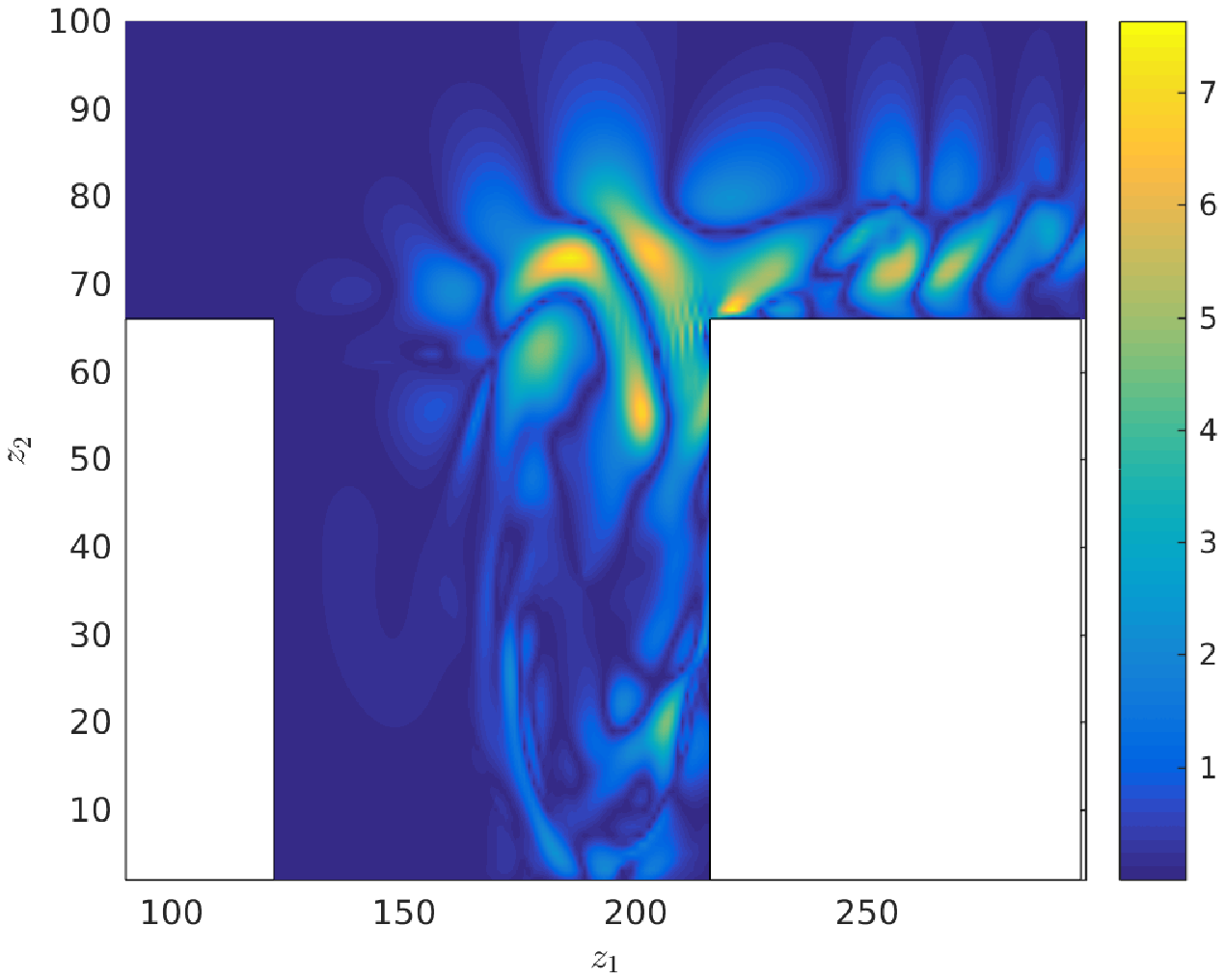}
        \caption{\GOBAL{} dictionary -- Mode 1.}
    \end{subfigure}%
    ~ \\
    \begin{subfigure}[t]{\the\numexpr\ratiolargeurthreecol \textwidth}
        \centering
        \includegraphics[width=\textwidth,angle=0]{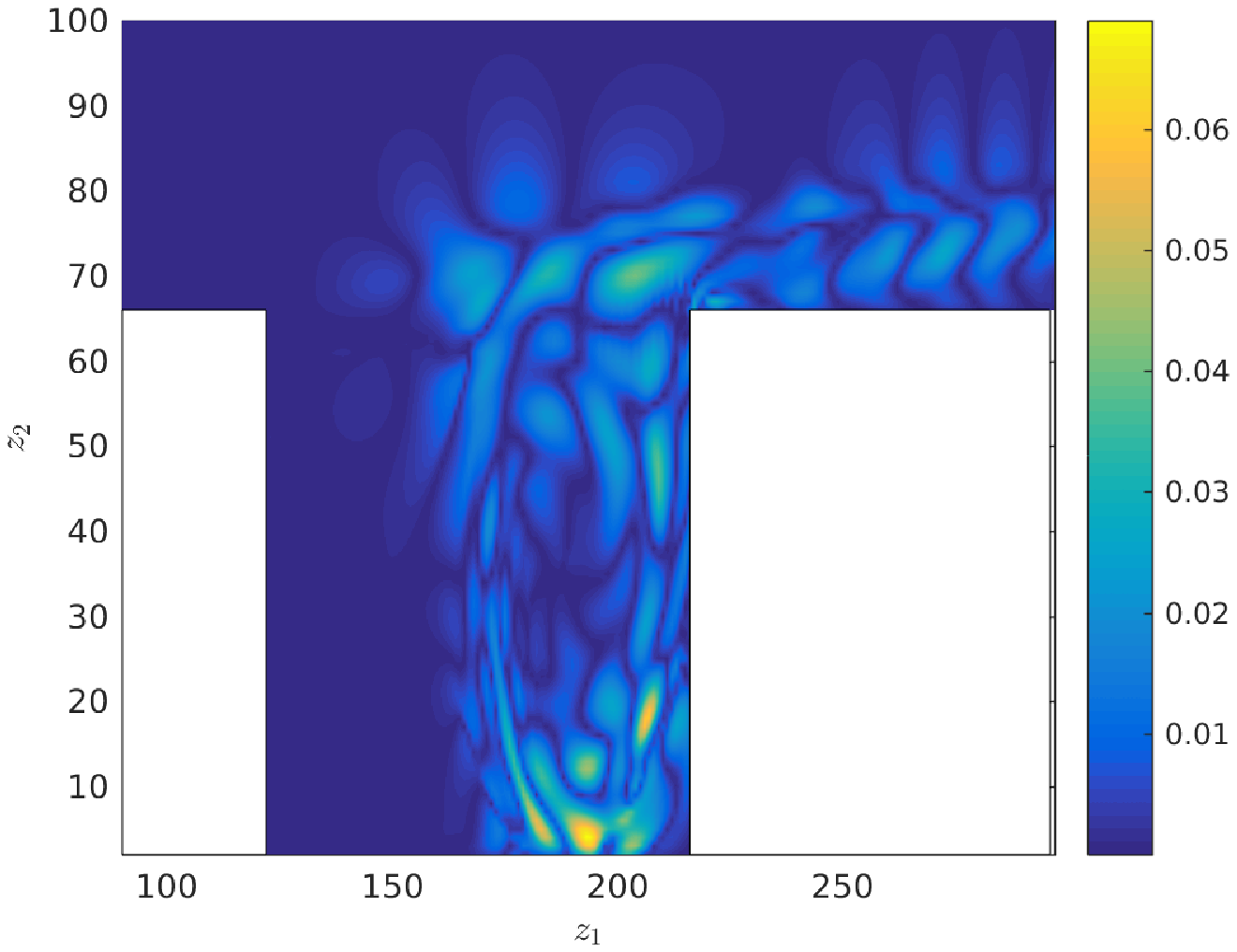}
        \caption{PCA dictionary -- Mode 10.}
    \end{subfigure}%
    \begin{subfigure}[t]{\the\numexpr\ratiolargeurthreecol \textwidth}
        \centering
        \includegraphics[width=\textwidth,angle=0]{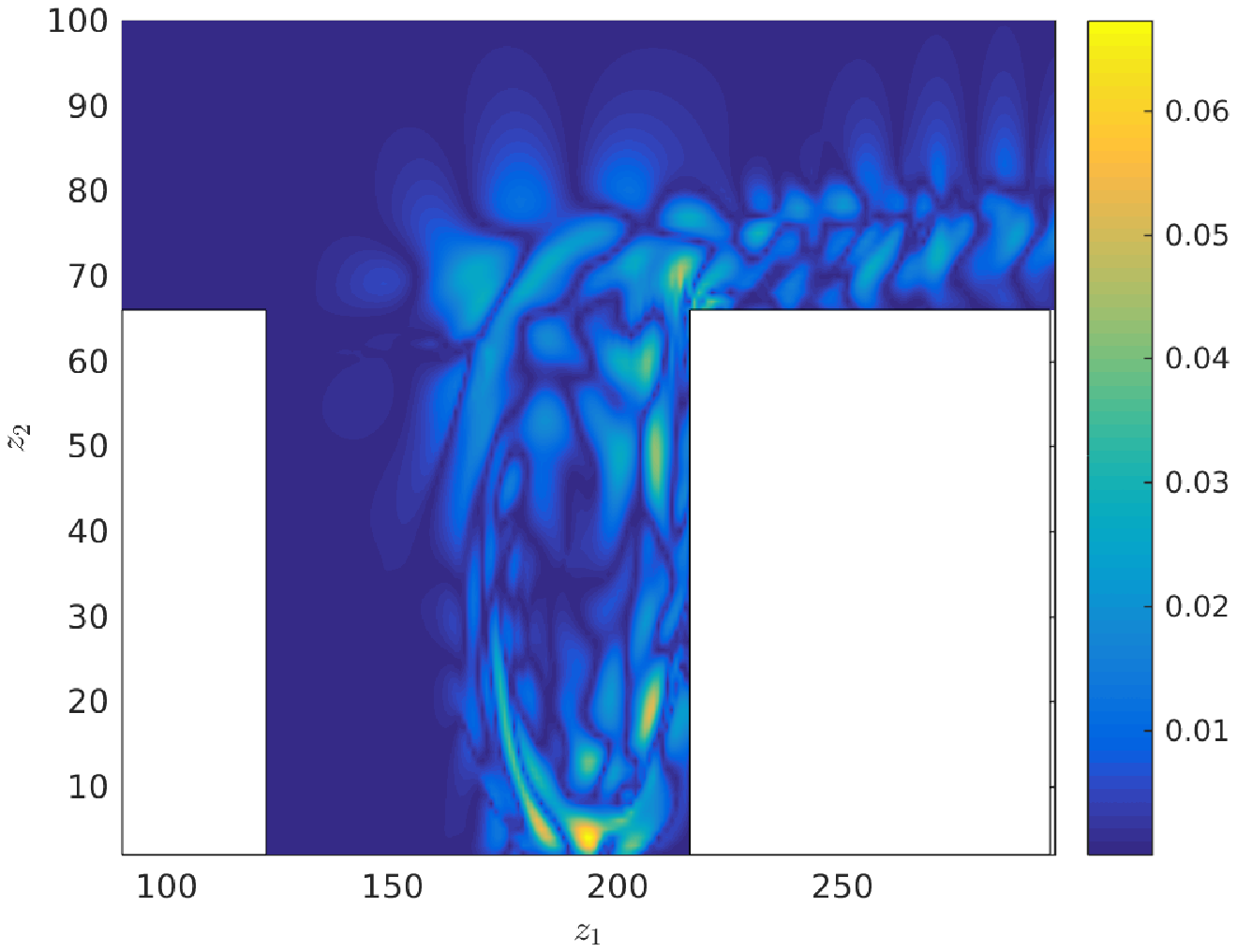}
        \caption{\KSVD{} dictionary -- Mode 10.}
    \end{subfigure}
    \begin{subfigure}[t]{\the\numexpr\ratiolargeurthreecol \textwidth}
        \centering
        \includegraphics[width=\textwidth,angle=0]{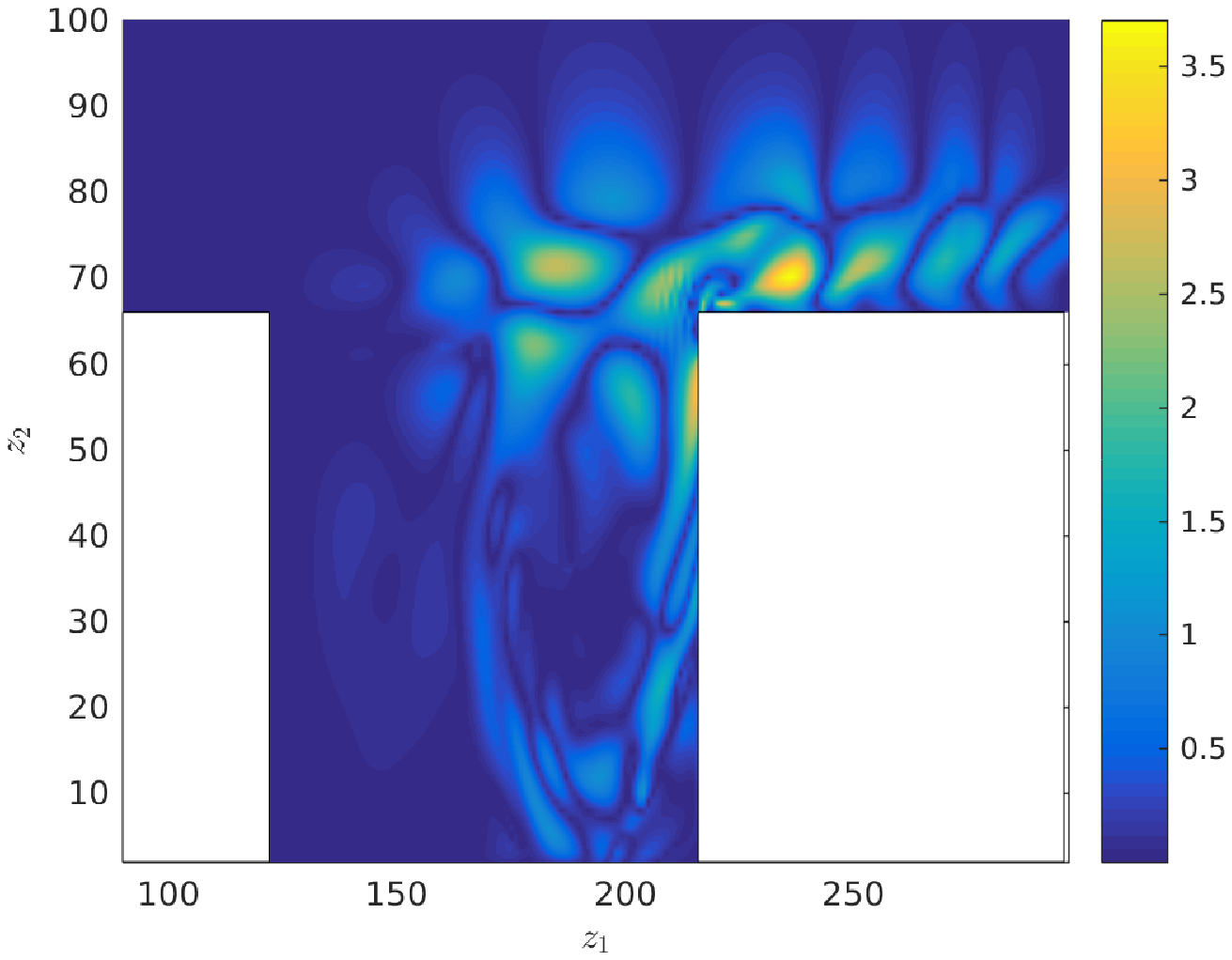}
        \caption{\GOBAL{} dictionary -- Mode 10.}
    \end{subfigure}%
    \caption{First (1st row) and 10-th (2nd row) modes of the different dictionaries. From left to right: PCA, \KSVD{} and \GOBAL{}.}
    \label{Fig_dico_modes}
\end{figure}

In our cavity flow simulation, the flow is horizontal, from left to right. Spatial oscillations develop in the shear layer above the cavity, while the left-most part of the domain essentially remains still and the flow varies very little about its mean $\vecymean$. A large-scale structure within the cavity as produced by the flow recirculation. One can refer to \cite{Rizi_etal_14} for a detailed discussion of the physics of this flow configuration.

The first and 10-th modes of the dictionaries from the different learning strategies are plotted in Fig.~\ref{Fig_dico_modes} for illustration. The main structures of the modes tend to reproduce the salient features of the flow. While the PCA and \KSVD{} dictionary look alike, at least for these two modes, the \GOBAL{}-based dictionary significantly departs from these two, while still exhibiting similar trends.

\subsection{Dependence on $\ndico$}	\label{dep_ndico}

A distinct feature of both the \KSVD{} and the present \GOBAL{} approaches is that they rely on \emph{redundant} approximations and allow to consistently inform a dictionary with more atoms than the number of sensors. At most $\nobs$ atoms are inferred from the data while the other atoms are associated with a vanishing coefficient. In contrast, the PCA-based approach informs the modes via Eq.~\eqref{data_misfit_linear} and, among the infinite number of solutions when $\ndico > \nobs$, selects the one with the least $\ldeux$-norm.

\begin{figure}[t!]
    \centering
    \includegraphics[height=\ratiolargeuronecol \textwidth,angle=-90]{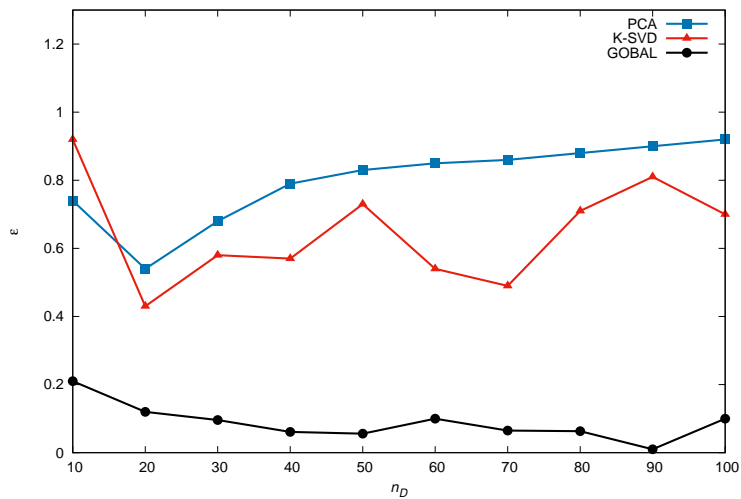}
    \caption{Influence of the size of the dictionary on the estimation error $\recerror$. $\nobs = 10$ sensors.}
    \label{Fig_sizedico}
\end{figure}

The recovery error $\recerr$ achieved by these different methods is plotted in Fig.~\ref{Fig_sizedico} as a function of the size $\ndico$ of the dictionary for $\nobs = 10$ sensors. When $\ndico = 10 = \nobs$, all dictionary modes can be informed from the available data. Yet, both the PCA- and the \KSVD{}-based approaches achieve a poor performance, with $\recerr \approx 0.8$. The conditioning of the observation of the dictionary atoms \via{} $\obsope \, \dicoCU$ is poor and prevents to accurately estimate $\matX^{\CV}$ from $\matS^{\CV}$. In this particular configuration, the sparsity constraint $\normLM{\vecx_i}{0} \le \nobs$ is always inactive and the sparse coding step, Eq.~\eqref{SC_CU_naive_X}, essentially reduces to the minimization of the data misfit \eqref{data_misfit_linear}. However, this minimization is achieved \via{} the greedy OMP algorithm and potentially deteriorates the recovery performance compared to \eqref{data_misfit_linear}. It results that, for $\ndico = \nobs$, the \KSVD{} approach performs worst. In contrast, the present \GOBAL{} approach enjoys a well conditioned predictor operator $\dico$ and accurately estimates $\matX^{\CV}$ and $\matY^{\CV}$, with $\recerr \approx 0.2$.

When the size of the dictionary grows beyond $\ndico = \nobs$, the performance of the PCA-based approach essentially gradually deteriorates as estimating $\vecx_i$ from $\vecsl{i}^{\CV}$ \via{} \eqref{data_misfit_linear} becomes increasingly ill-posed. In contrast, the performance of the \KSVD{}-based approach remains roughly constant, $\recerr \approx 0.7$. However, since $\dicoCU$ stems from $\matY$ solely, observations $\vecsl{i}^{\CV}$ do not necessarily admit a sparse representation in the linear span of the columns of $\obsope \, \dicoCU$ and the estimation of the coefficients \via{} the sparse coding step \eqref{SC_CU_naive_X} is likely to lead to a poor set of coefficients, hence a poor $\recerr$. The \GOBAL{} approach, however, precisely addresses this issue and explicitly determines the predictor operator such that $\matS^{\CV}$ is likely to admit a compressible representation in $\spanLM \left\{\dicol{l}\right\}_l$. The performance is seen to be good and to (slightly) improve when the size of the dictionary increases, allowing for better tailored atoms to be used for reconstructing a given field $\vecy^{\CV}$ from observations $\vecs^{\CV}$.

\begin{figure}[t!]
    \centering
    \begin{subfigure}[t]{\the\numexpr\ratiolargeurtwocol \textwidth}
        \centering
        \includegraphics[width=\textwidth,angle=0]{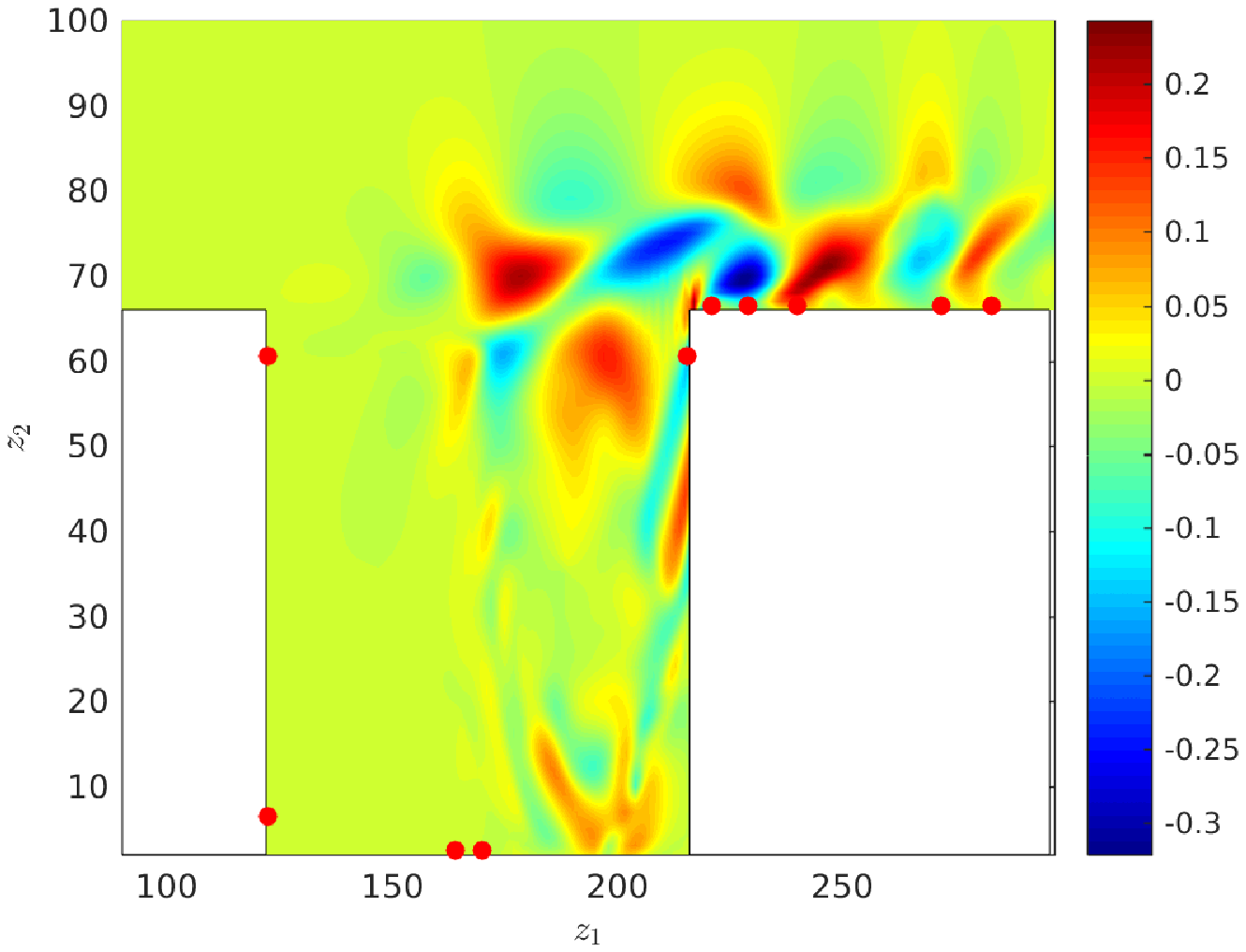}
        \caption{Truth.}
    \end{subfigure}%
    \begin{subfigure}[t]{\the\numexpr\ratiolargeurtwocol \textwidth}
        \centering
        \includegraphics[width=\textwidth,angle=0]{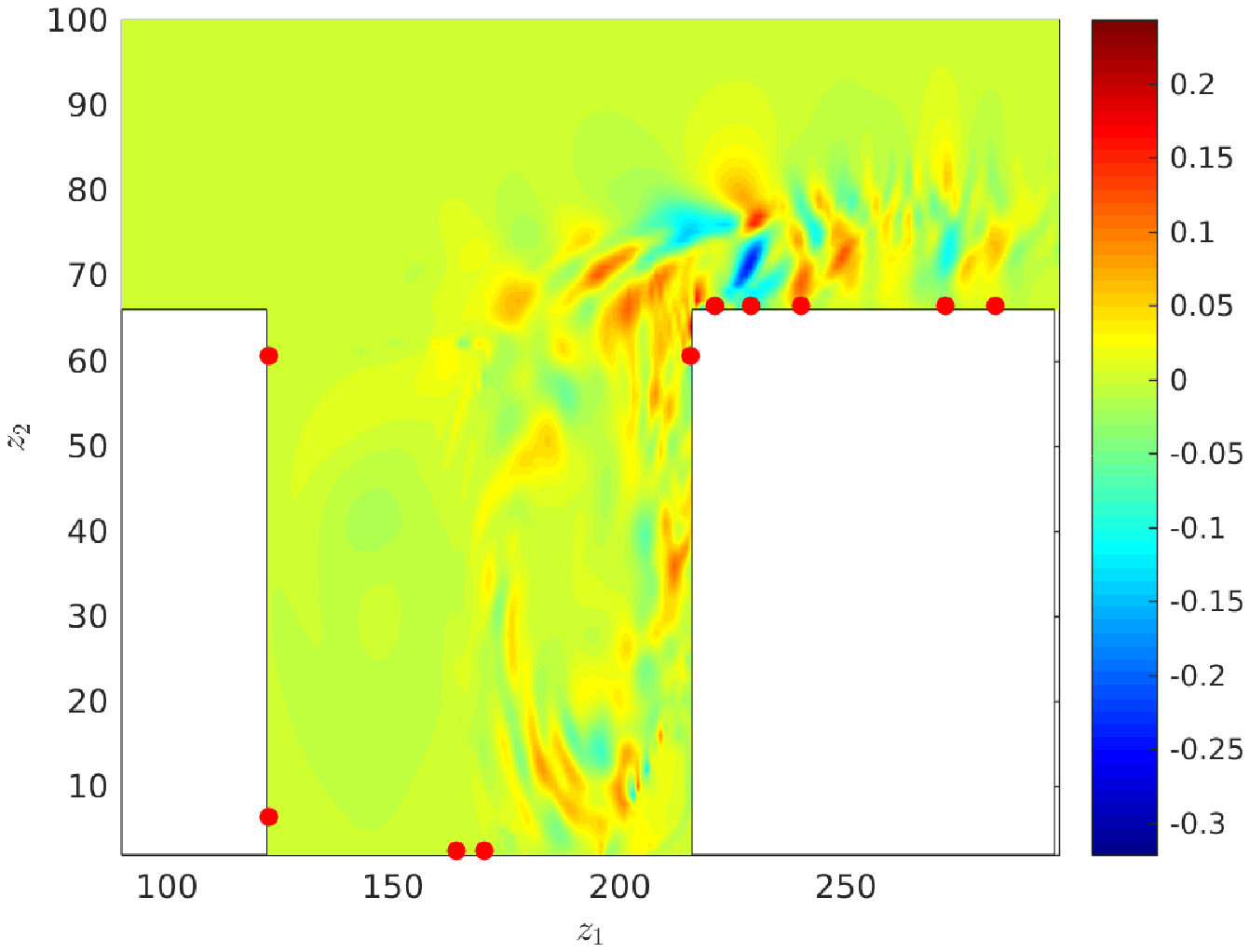}
        \caption{PCA.}
    \end{subfigure}
    ~ \\
    \begin{subfigure}[t]{\the\numexpr\ratiolargeurtwocol \textwidth}
        \centering
        \includegraphics[width=\textwidth,angle=0]{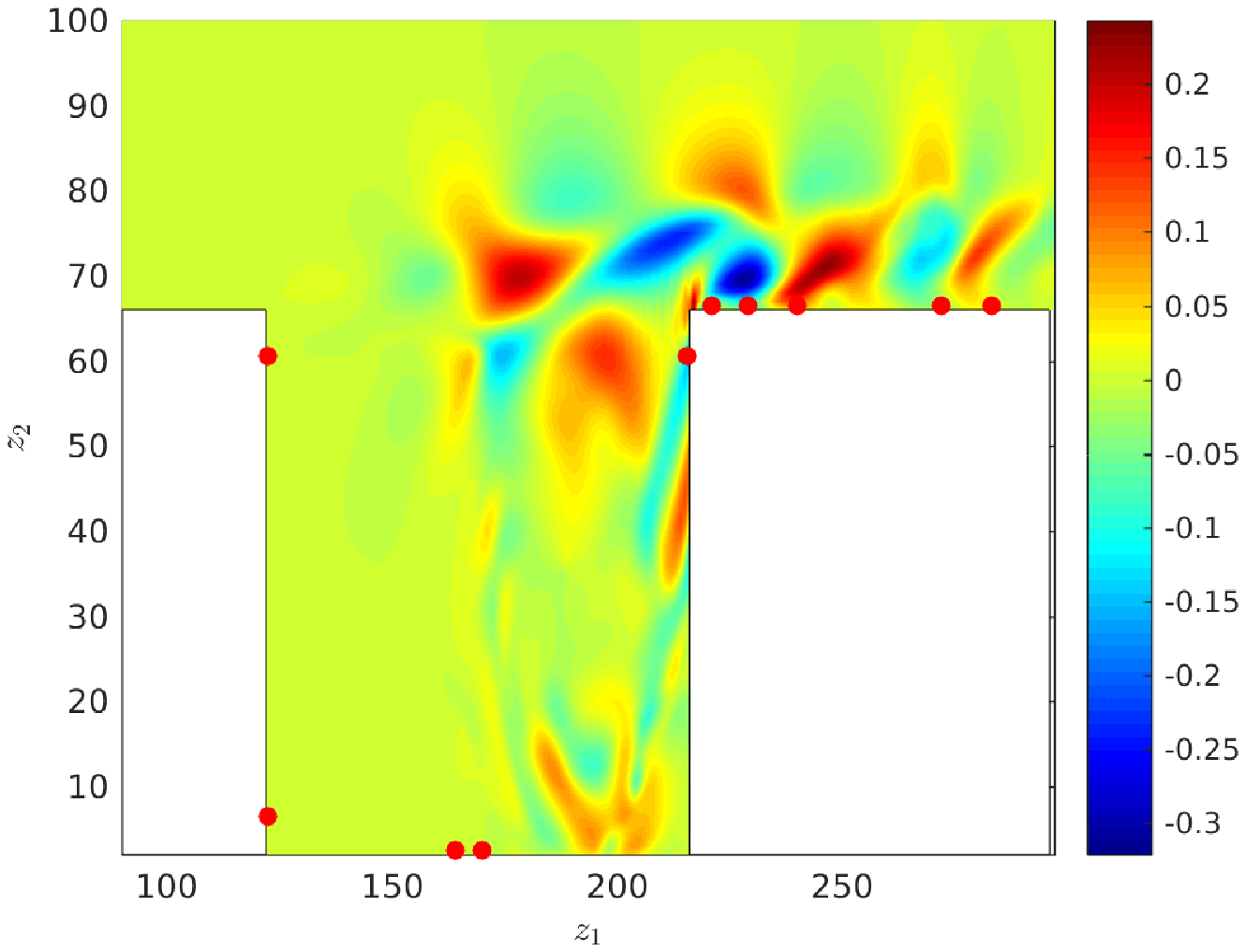}        
        \caption{\GOBAL{}.}
    \end{subfigure}%
    \begin{subfigure}[t]{\the\numexpr\ratiolargeurtwocol \textwidth}
        \centering
        \includegraphics[width=\textwidth,angle=0]{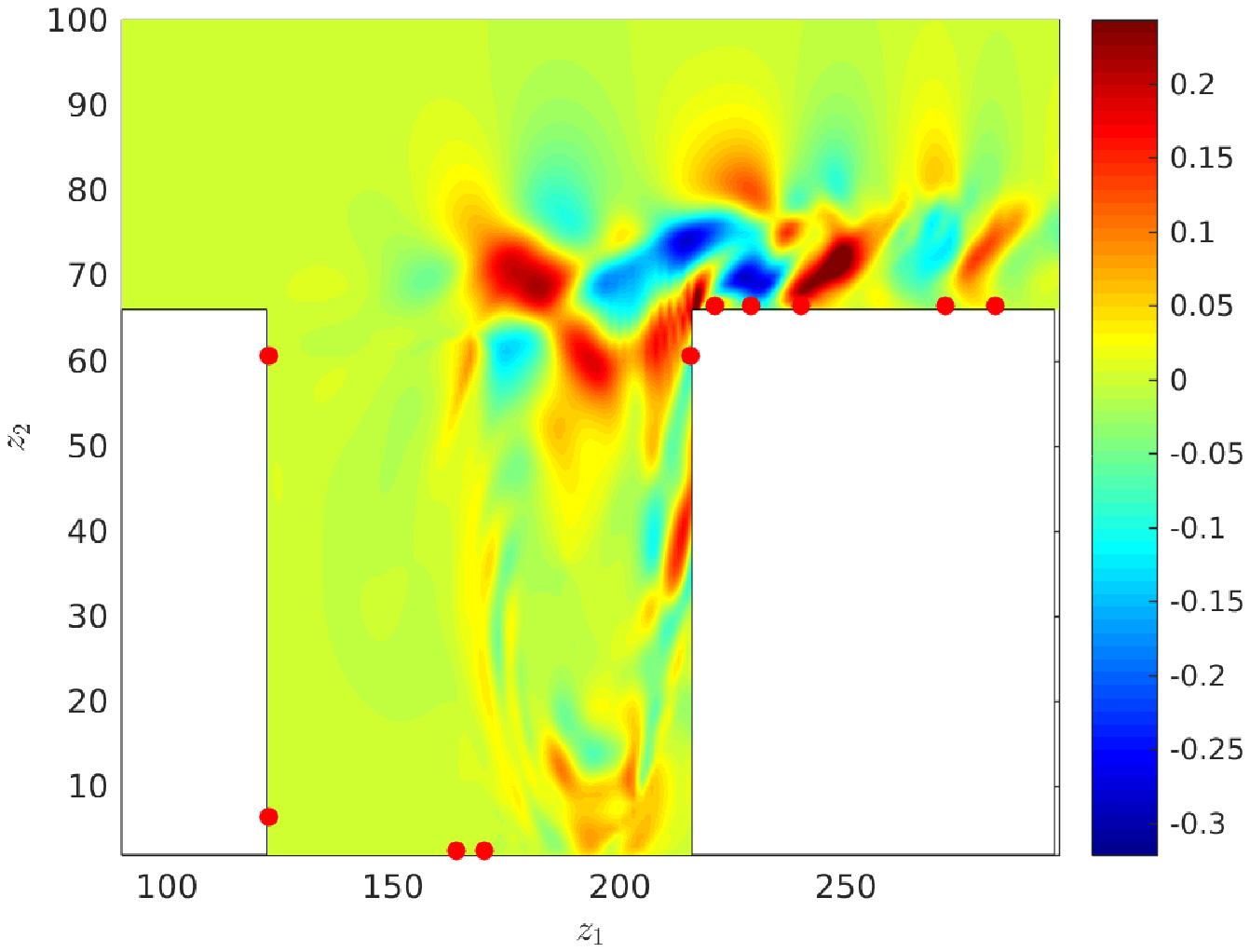}
        \caption{\KSVD{}.}
    \end{subfigure}
    \caption{Example of field estimation using different dictionary learning strategies. The truth is in the top left and the sensors are denoted with large red dots. Plotted are the fluctuations of the flow field from its (empirical) mean as estimated from the training set.}
    \label{Fig_recovery_fields}
\end{figure}

The recovery performance is also illustrated in Fig.~\ref{Fig_recovery_fields} where the different methods are employed to infer a field chosen at random from $\left\{\vecyl{i}^{\CV}\right\}_i$ from associated observations, as given by the shear stress measurement at the locations of the sensors. The sensors are denoted with red dots in the figure. The recovered field is plotted for the PCA- (top right), \KSVD- (bottom right) and \GOBAL{}-based (bottom left) methods, along with the actual field (top left). The superior performance of the \GOBAL{} approach is clearly visible.

\subsection{Dependence on $\nobs$}	\label{dep_nobs}

The quality and quantity of information is pivotal to a reliable estimation. The impact of the quantity is here studied in terms of number of sensors. From the 10 sensors considered in the above section, the different recovery methods are now applied only relying on the first $1 \le \nobs \le 10$ sensors from the full set of 10 sensors. Note that the set of 10 sensors considered so far was determined, from a large set of sensor placement realizations, as that giving the best recovery performance for the \KSVD{}-based approach. Within this ``best set'', they are not sorted or ranked in any way.

\begin{figure}[t!]
    \centering
    \begin{subfigure}[t]{\the\numexpr\ratiolargeurtwocol \textwidth}
        \centering
        \includegraphics[height=\textwidth,angle=-90]{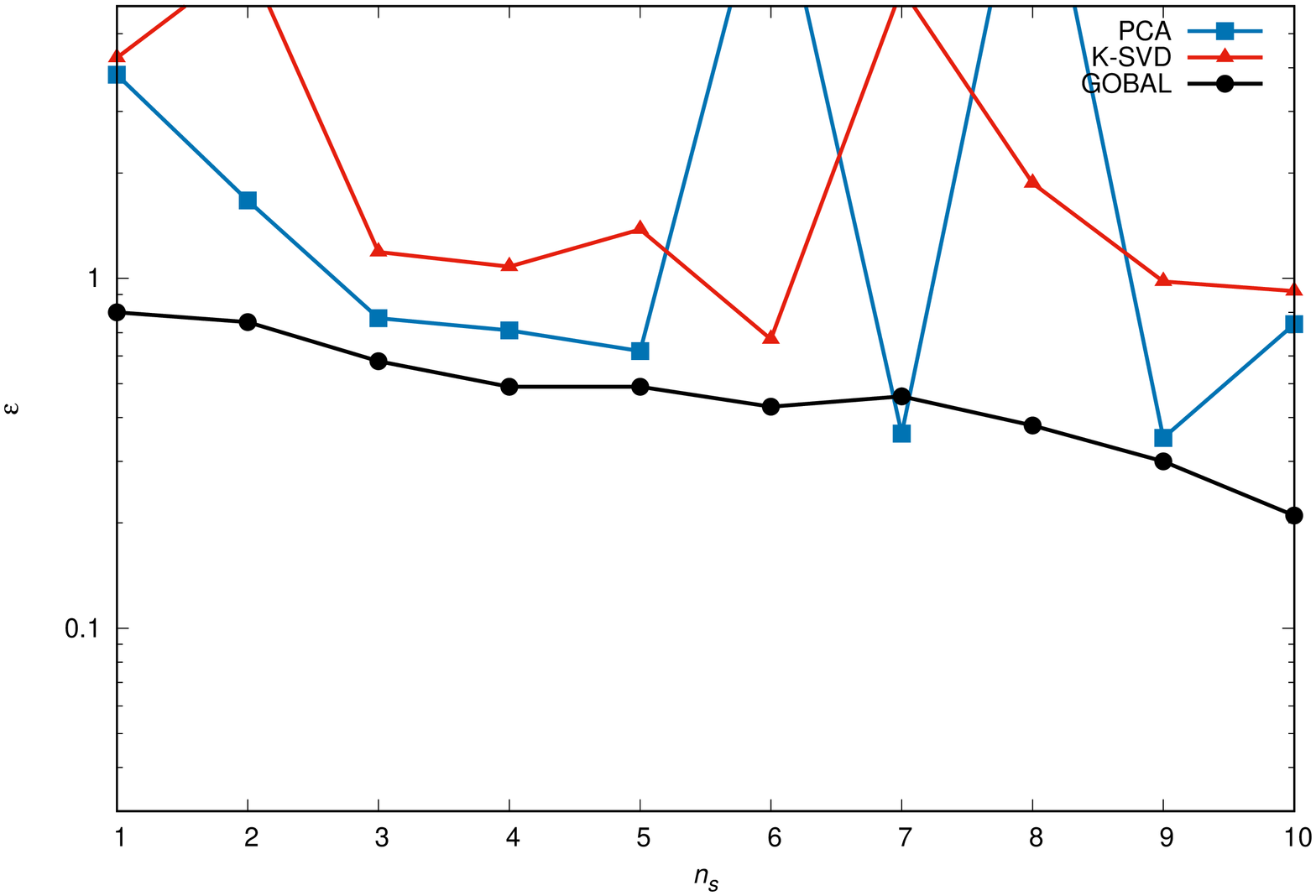}
        \caption{$\ndico = 1 \times \nobs$.}
    \end{subfigure}%
    \begin{subfigure}[t]{\the\numexpr\ratiolargeurtwocol \textwidth}
        \centering
        \includegraphics[height=\textwidth,angle=-90]{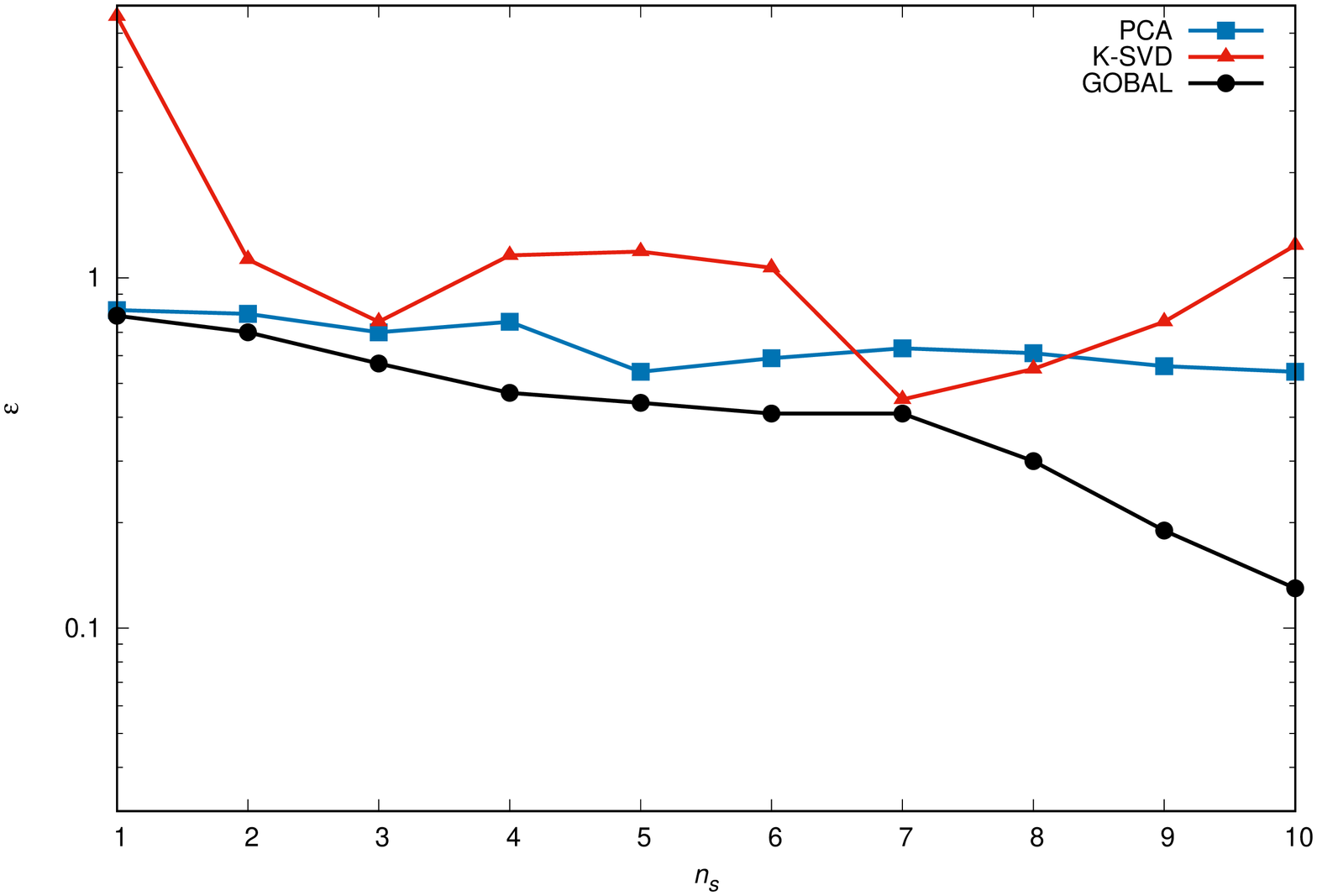}
        \caption{$\ndico = 2 \times \nobs$.}
    \end{subfigure}
    ~ \\
    \begin{subfigure}[t]{\the\numexpr\ratiolargeurtwocol \textwidth}
        \centering
        \includegraphics[height=\textwidth,angle=-90]{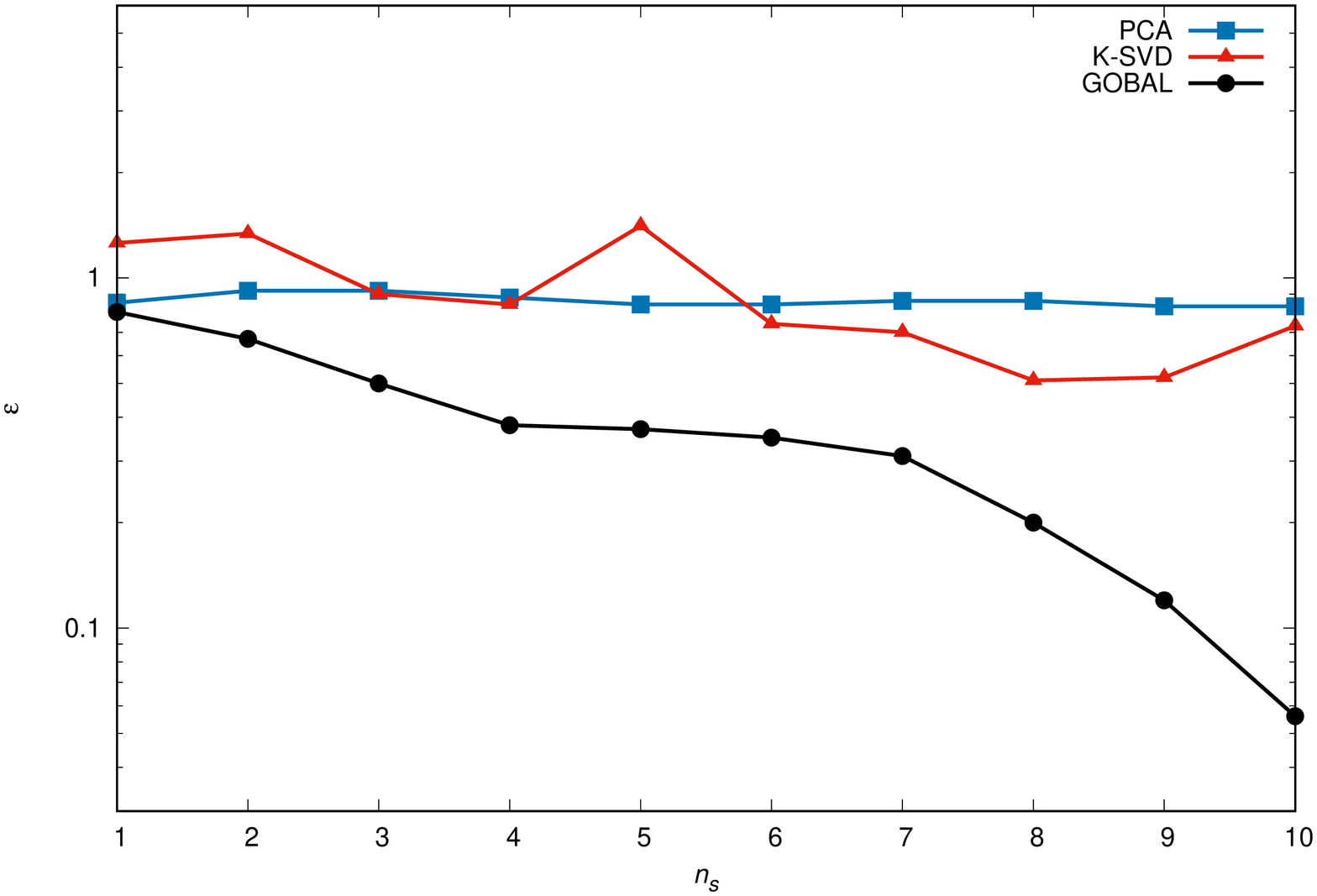}
        \caption{$\ndico = 5 \times \nobs$.}
    \end{subfigure}%
    \begin{subfigure}[t]{\the\numexpr\ratiolargeurtwocol \textwidth}
        \centering
        \includegraphics[height=\textwidth,angle=-90]{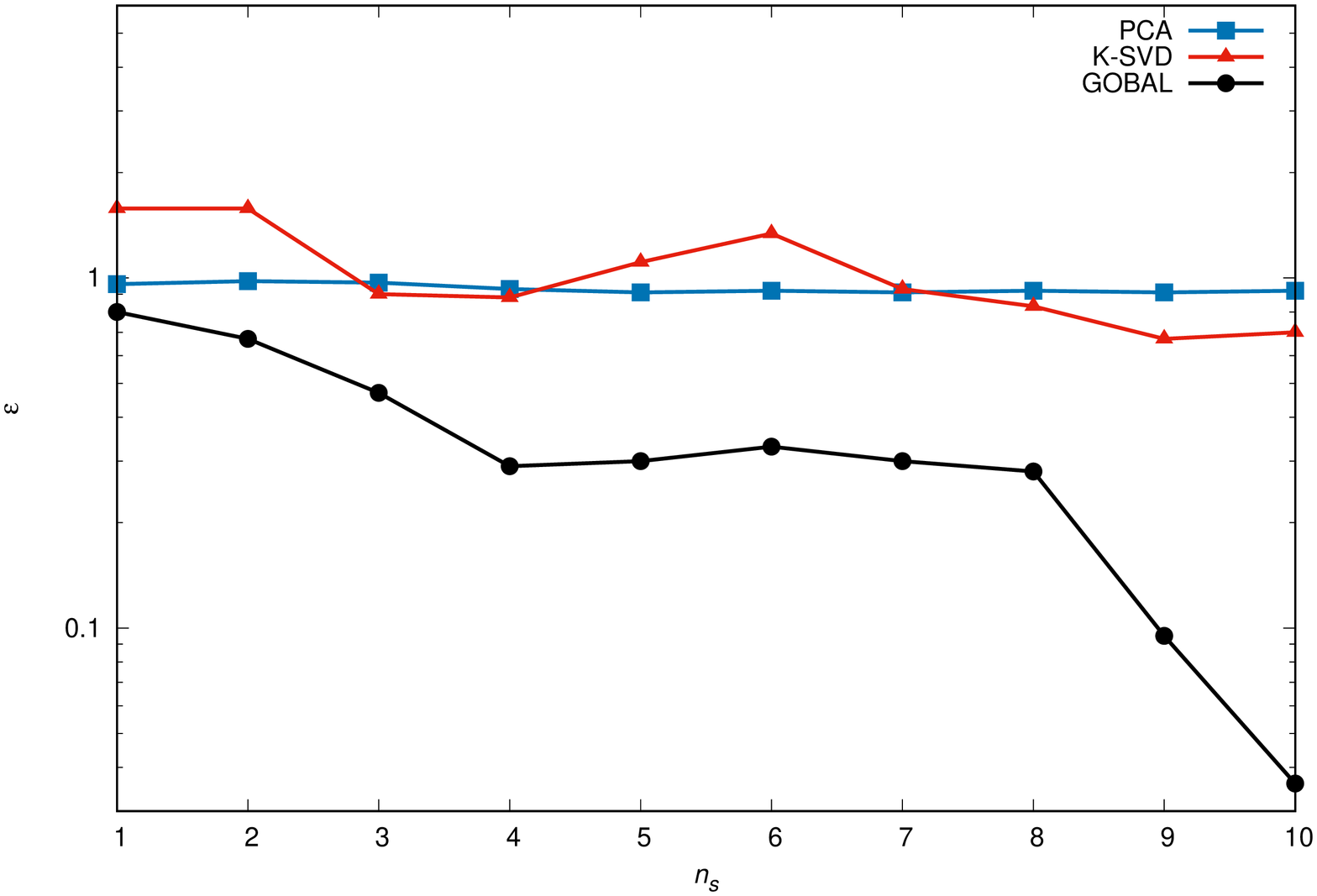}
        \caption{$\ndico = 10 \times \nobs$.}
    \end{subfigure}
    \caption{Influence of the number of sensors on the estimation error $\recerror$ for several relative size of the dictionary.}
    \label{Fig_coefmult}
\end{figure}

The influence of the number of sensors on the recovery accuracy is illustrated in Fig.~\ref{Fig_coefmult} for different sizes of dictionary. It is important to note that the size of the dictionary here increases with the number of sensors, keeping the ratio $m = \ndico \slash \nobs$ constant.
For a given dictionary, the performance of the \GOBAL{} approach is seen to significantly improve when more information becomes available (increasing $\nobs$). In contrast, the performance of the \KSVD{}-based approach hardly improves and remains rather poor, while the PCA-based performance does not improve and remains around a relative error $\recerr \approx 1$.
Again, the \GOBAL{} method leads to a situation where the observations are likely to admit a sparse representation in the linear span of the predictor operator, hence allowing an accurate recovery, while no such situation occurs with the other two approaches. It is remarkable that this superior performance of \GOBAL{} is achieved despite the fact that the sensors are located to be ``optimal'' for \emph{another} method, namely \KSVD{}.


\subsection{Robustness}	\label{Sec_robustness}

An important point in most, if not all, numerical methods is robustness. To be applicable in practice, solution methods should gracefully deteriorate the solution accuracy when input data are affected by a weak deviation from their nominal values. These deviations may come from round-off and truncation errors in the numerics. However, in the present context, the largest (by far) source of error, also termed noise, is expected to originate from the sensor measurements. Characterizing the influence of measurement noise onto the solution accuracy is necessary to demonstrate the benefit and performance of our method.

The noise is assumed to be a Gaussian vector with zero-mean and diagonal variance matrix, as discussed in Sec.~\ref{SBL_proba_est}, $\vecnoise \sim \gauss\left(\veczero, \noisevarmult\right)$. Our \GOBAL{} method determines the posterior density of the estimate under an \emph{additive} noise assumption, see Eq.~\eqref{observation_model}. However, we here test our method in a context \emph{different} from the one it is derived for: the accuracy of the solution is studied as the variance $\noisevarmult$ of a \emph{multiplicative} measurement noise varies, that is, the measurement data are actually given by
\begin{align}
\vecs = \left(\boldsymbol{1}_{\nobs} + \gauss\left(\veczero, \noisevarmult\right)\right) \odot \dico \, \vecx, \label{observation_model_actual}
\end{align}
with $\odot$ the Hadamard product. Noise affects both the training data and the \textit{in situ} measurements and a given QoI is then associated with \emph{several} measurements. The \GOBAL{} algorithm is then run on an extended training set $\left\{\matY, \left\{\matS\left(\stovarnoise^{(j)}\right)\right\}_j\right\}$. In this work, five realizations, $j = 1, \ldots, 5$, of each measurement vector are considered.

\begin{figure}[t!]
    \centering
    \includegraphics[height=\ratiolargeuronecol \textwidth,angle=-90]{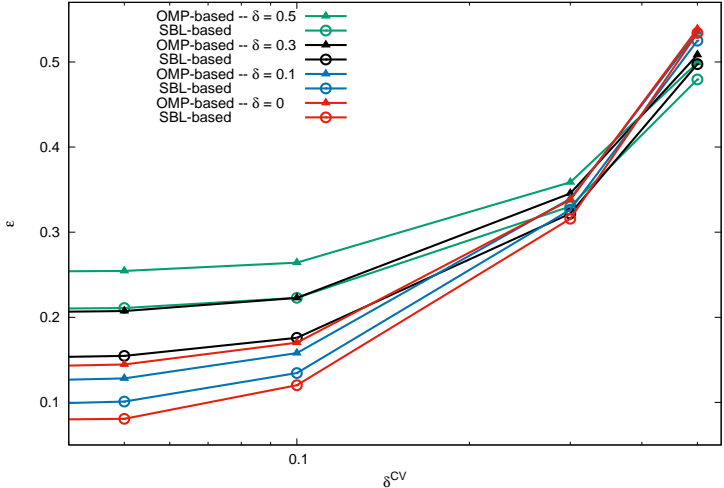}
    \caption{Influence of the noise in the training and online measurements on the recovery accuracy. $\nobs = 10$, $\ndico = 50$. Best seen in color (online version).}
    \label{Fig_noise}
\end{figure}

The Figure~\ref{Fig_noise} shows the evolution of the recovery accuracy $\recerr$ as a function of the measurement noise standard deviation $\deltanoise$ in the training set $\matS$, and the noise standard deviation $\deltanoise^{\CV}$ in the actual measurements $\matS^{\CV}$. Several comments are in order. The accuracy of the recovery decreases when the noise variance $\deltanoise^{\CV}$ increases. However, when the noise is strong ($\deltanoise^{\CV} = 0.5$), the best performance is achieved with dictionaries trained with strong noise as well, $\deltanoise = 0.3$ and $\deltanoise = 0.5$. If properly trained, the dictionaries learned from noisy data are hence more robust to noise in the measurements. However, this is at the price of a reduced performance if the measurement noise is low, in which case dictionaries trained without noise perform best.
This shows how the learning can be made robust to the expected level of noise of \textit{in situ} data and that the best performance is obtained when noise in the training step is as similar to that actually affecting the sensors as possible.

Another observation is that, for any noise level, the SBL-based version of \GOBAL{} always performs significantly better than its OMP-based counterpart. There are two reasons for this. First, the OMP-variant of \GOBAL{} is here used with $\Ksparsity = \nobs$, not allowing any redundancy in the data to inform the non-vanishing mode coefficients, and hence being prone to inaccuracy. Second, OMP is a greedy approach which converges to a local minimum. To address this issue, several initial conditions are usually used, and the best solution is retained. However, in practical situations as mimicked here, the ``best'' solution cannot be known, nor \aposteriori{} verified. All solutions from the OMP step are hence equivalent, with no mean of discriminating the ``good'' from the ``poor'' solutions. We hence use a \emph{unique} initial condition for OMP, identical for the training and the online step, with the risk for the OMP-based SC step to reach a poor local minimum. The Sparse Bayesian Learning variant of our SC step does not suffer these two limitations and potentially achieves a better overall recovery performance.


\subsection{Probabilistic estimate} \label{Sec_posterior}

\begin{figure}[t!]
    \centering
    \begin{subfigure}[t]{\the\numexpr\ratiolargeurtwocol \textwidth}
        \centering
        \includegraphics[width=\textwidth,angle=0]{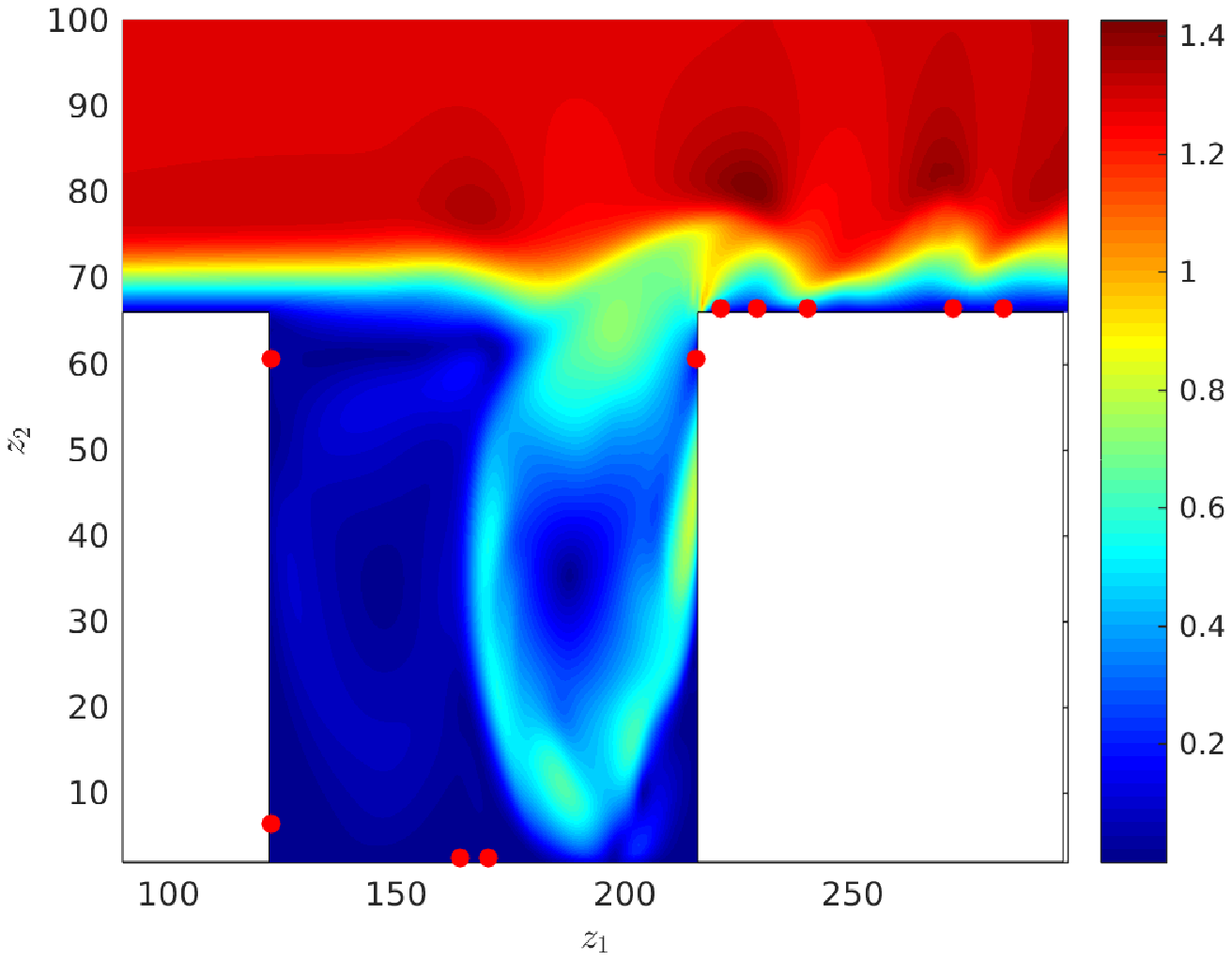}
        \caption{Truth, $\vecy$.}
    \end{subfigure}%
    \begin{subfigure}[t]{\the\numexpr\ratiolargeurtwocol \textwidth}
        \centering
        \includegraphics[width=\textwidth,angle=0]{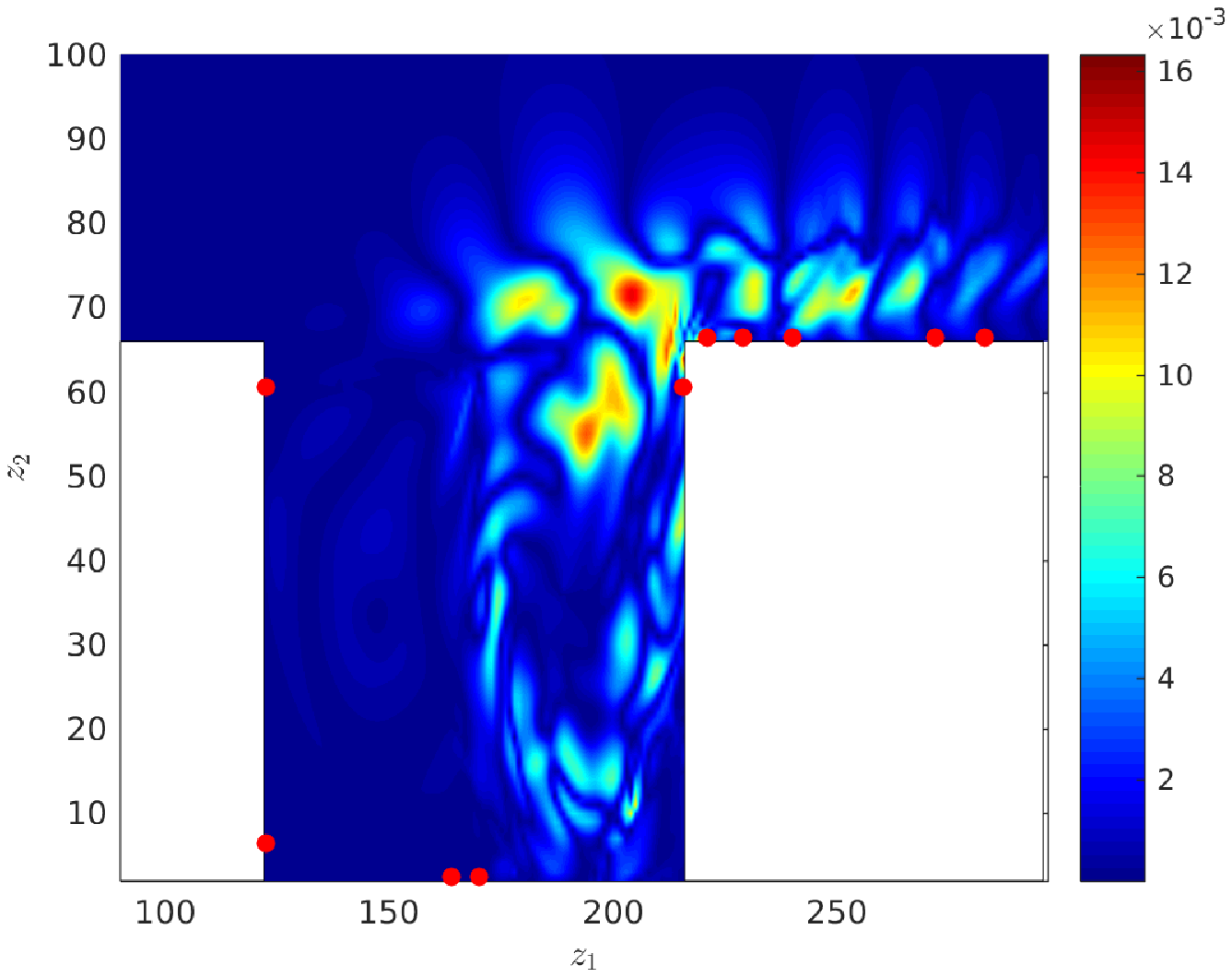}
        \caption{Absolute value of the \GOBAL{}-based reconstruction error, $\abs{\vecy - \vecyest}$.}
    \end{subfigure}
    ~ \\
    \begin{subfigure}[t]{\the\numexpr\ratiolargeurtwocol \textwidth}
        \centering
        \includegraphics[width=\textwidth,angle=0]{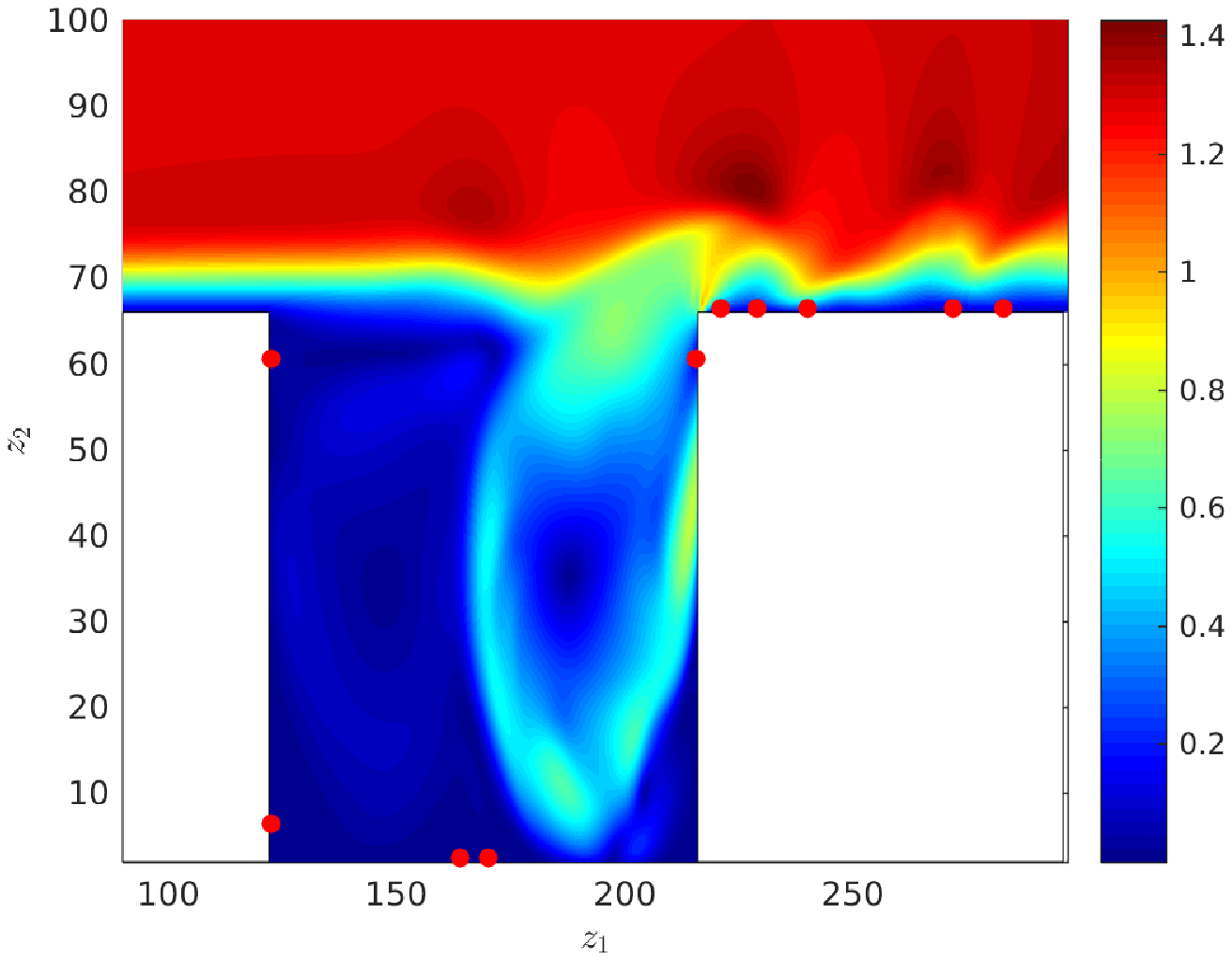}        
        \caption{\GOBAL{} -- Posterior mean field, $\vecyest = \dicoCU \, \postmean$.}
    \end{subfigure}%
    \begin{subfigure}[t]{\the\numexpr\ratiolargeurtwocol \textwidth}
        \centering
        \includegraphics[width=\textwidth,angle=0]{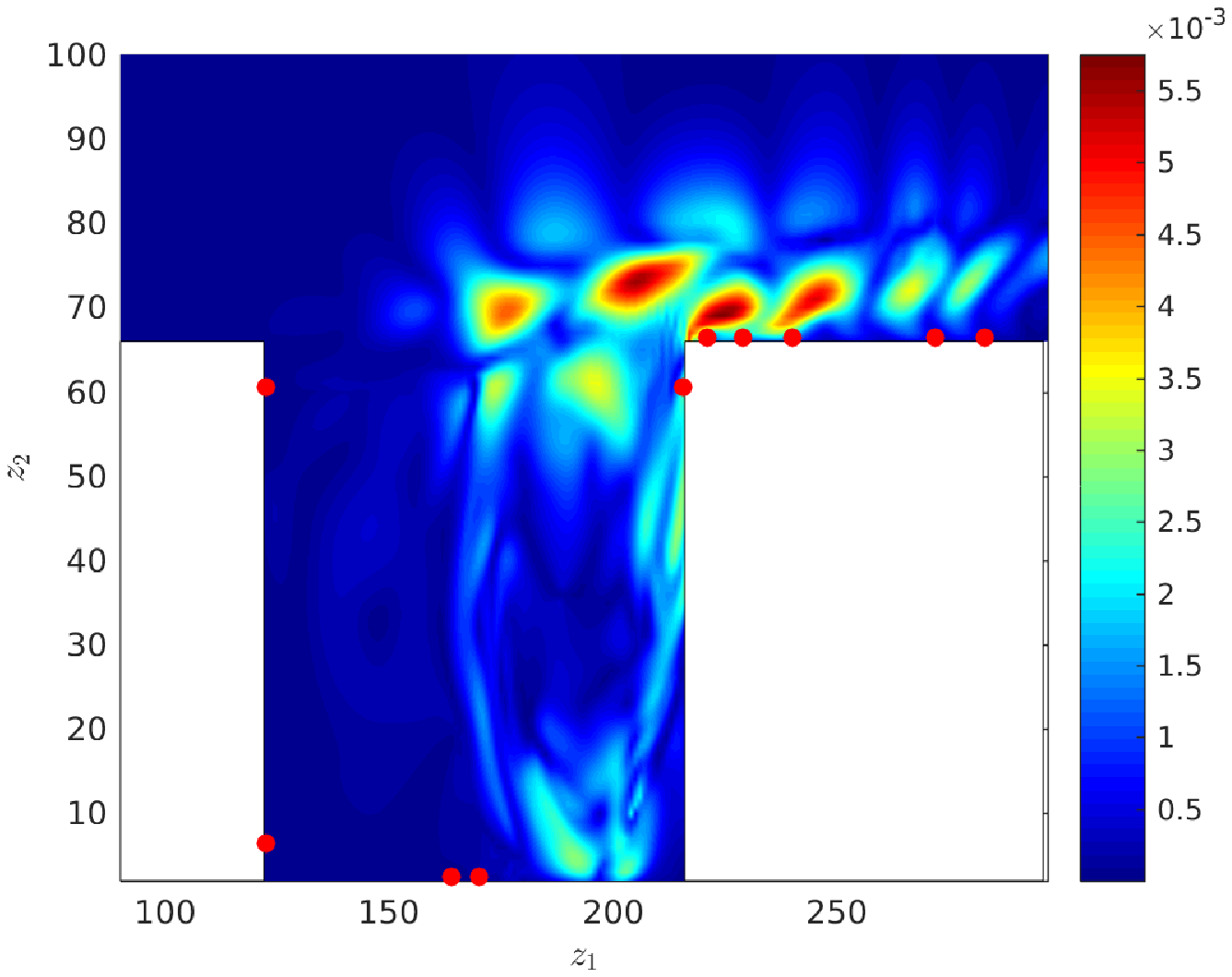}
        \caption{\GOBAL{} -- Posterior standard deviation, $\left[\text{diag}\left(\dicoCU \, \postvariance \, \dicoCU^\transpose\right)\right]^{1/2}$.}
    \end{subfigure}
    \caption{Probabilistic estimation with \GOBAL{}. The truth is in the top left.}
    \label{Fig_recovery_mean_std}
\end{figure}

The probabilistic estimate provided by our \GOBAL{} method is illustrated in Fig.~\ref{Fig_recovery_mean_std}. The posterior mean field, $\vecyest = \dicoCU \, \postmean$ (bottom left plot) is seen to compare well with the truth (top left plot). The diagonal part of the posterior standard deviation, $\left[\text{diag}\left(\dicoCU \, \postvariance \, \dicoCU^\transpose\right)\right]^{1/2}$, (bottom right plot) illustrates the spatial regions where noise in the measurements, and more generally the limitation introduced by the likelihood model, leads to uncertainty in the estimate. Clearly, regions with the largest estimated standard deviation are those where the flow exhibits the most fluctuations, \ie{}, the downstream part of the shear layer and the main recirculation vortex within the cavity. In the top right plot, the (absolute value) reconstruction error of the posterior mean, $\abs{\vecy - \vecyest}$, also exhibits the same general pattern as the posterior standard deviation. That is, regions of the estimated field identified as the most uncertain are indeed roughly those where the error is the largest. It is however important to note that the error is low, even its maximum $\normLM{\vecy - \vecyest}{\infty} \approx 1.6 \, 10^{-2}$, compared to the field itself which lies between 0 and about 1.4. This is remarkable since the estimation method targets a reconstruction in the $\ldeux$-sense rather than in the $\linf$-sense.

\section{Closing remarks}	\label{Sec_Conclusion}

In this paper, an estimation technique for inferring high-dimensional quantities is introduced. It relies on learning an approximation basis, the dictionary, in which the quantity of interest is estimated from a set of observations by minimizing the data misfit. The original contribution of the present work is that the dictionary is learned with the specific constraint that it has to be \emph{observable} from the available sensors. This makes estimation of the associated coefficients more accurate and reliable. An algorithm is presented and allows to derive an observable dictionary from a training set of observations and fields to be inferred. Estimation in the online phase, when \textit{in situ}, is achieved via an algorithm combining Bayesian estimation and sparse recovery. It results in a probabilistic estimation of the quantity to be inferred. The formulation of the proposed recovery procedure does not involve the size of the high-dimensional quantity of interest but instead scales with the size of the training set, usually much smaller. It results in computationally efficient learning.
The method is illustrated in the case of the two-dimensional flow over an open cavity whose velocity field is to be estimated from a small set of point sensors located on the wall. Comparisons with established PCA- and K-SVD-based estimation methods show the superior performance of the present approach.

This study clearly emphasizes the pivotal role of the observations quality onto the recovery performance. While sensor placement is a well-known aspect, the need for a \emph{joint} derivation of the estimation procedure and determination of an approximation basis is clearly demonstrated in the present work. We strongly advocate learning \emph{both} the representation basis and the recovery procedure. While not done here, sensor placement could, and actually should, be learned at the same time, as it strongly impacts the resulting observability properties of the dictionary. An integrated approach, accounting for all these aspects as early in the estimation procedure as possible, is key to an efficient and reliable method.

The present work is a first step towards a more general framework. In addition to coupling dictionary learning with sensor placement, future extensions include nonlinear observers where the measurements are not well approximated in the image space of a linear predictor operator. Derivation of the learning step in this more general framework leads to a bi-level optimization problem and will be presented in a future paper.

The estimation procedure introduced here relies on the minimization of the reconstruction residual $\ldeux$-norm. While a common approach, this choice is questionable in some situations. For instance, while suitable for minimizing the estimation error in amplitude, this is a poor choice for minimizing the error in \emph{phase}. As an example, consider a spatially-located feature well estimated but suffering a slight spatial shift. The associated $\ldeux$-norm estimation error would then be large while the recovery is ``good'' in the eyeball norm. This motivates alternative norms to be used. Among other choices, Wasserstein distances are commonly used in some other contexts, \eg{}, image processing, and may be a more appropriate choice. To alleviate the significant computational overhead arising from working with Wasserstein distances, one could instead rely on a framework stable to deformations as, say, under a scattering transform preserving some invariants, to some extent, \cite{Scattering_transform_Mallat}.

More generally, an efficient estimation procedure first learns a suitable transformation such that, in the resulting framework, the inverse problem formulates as a problem easier to solve than its original counterpart. Here, transformations promoting sparse solutions where exploited, allowing efficient sparse recovery tools to be used. Reformulating the estimation problem in a reproducing kernel Hilbert space (RKHS), with a suitable kernel and inner product learned from the data, is a more general approach and may allow to ``untangle'' a complex relationship between, say observations and approximation coefficients, paving the way for an efficient use of tools from the world of linear methods. Learning the manifold onto which the data lie, and a suitable kernel, allows application of techniques presented in this work to a more general context and provides new perspectives for inverse problems. Restricting to only one example, the current work may readily be extended into a \emph{filter} for observing dynamical systems. Relying on \emph{extended} observations, as collected over a given time-horizon, the present observable dictionary learning method can be reformulated in a suitable RKHS learned from the data. Since the estimation problem then relies on an extended set of observations, we expect superior performance as compared to the present static estimator.

These developments are the subject of on-going efforts.

\section*{Acknowledgement}
The first author (LM) gratefully acknowledges stimulating discussions with Youssef M Marzouk and Alessio Spantini.

\section*{Conflict of interest}
On behalf of all authors, the corresponding author states that there is no conflict of interest. 

\bibliographystyle{lionel_style}	
\bibliography{biblio}

\appendix

\end{document}